\documentclass[acmtog,nonacm]{acmart}

\AtBeginDocument{%
  }

\acmJournal{TOG}

\usepackage[capitalize]{cleveref}
\crefname{section}{Sec.}{Secs.}
\Crefname{section}{Section}{Sections}
\Crefname{table}{Table}{Tables}
\crefname{table}{Tab.}{Tabs.}

\usepackage{bm}

\usepackage{multirow}

\usepackage{pifont}
\newcommand{\cmark}{\ding{51}}%
\newcommand{\xmark}{\ding{55}}%

\usepackage{tablefootnote}

\usepackage{soul,color}
\definecolor{Red}{cmyk}{0,1,1,0}
\definecolor{Green}{cmyk}{1,0,1,0}
\definecolor{Cyan}{cmyk}{1,0,0,0}
\definecolor{Purple}{cmyk}{0.45,0.86,0,0}
\definecolor{Rosolic}{cmyk}{0.00,1.00,0.50,0}
\definecolor{Blue}{cmyk}{1.00,1.00,0.00,0}
\definecolor{Orange}{cmyk}{0,0.52,0.80,0}
\definecolor{Black}{cmyk}{1,0,0,1}
\definecolor{light-blue}{rgb}{0.8,0.85,1}
\definecolor{teaser-green}{rgb}{0.35,0.5,0.25}

\begin{document}

\title{VMesh: Hybrid Volume-Mesh Representation for Efficient View Synthesis}

\author{Yuan-Chen Guo}
\affiliation{%
  \institution{Tsinghua University}
  \state{Beijing}
  \country{China}}
\email{guoyc19@mails.tsinghua.edu.cn} 

\author{Yan-Pei Cao}
\affiliation{%
  \institution{ARC Lab, Tencent PCG}
  \state{Beijing}
  \country{China}}
\email{caoyanpei@gmail.com}  

\author{Chen Wang}
\affiliation{%
  \institution{Tsinghua University}
  \state{Beijing}
  \country{China}}
\email{wchen20@mails.tsinghua.edu.cn}  

\author{Yu He}
\affiliation{%
  \institution{BIMSA}
  \state{Beijing}
  \country{China}}
\email{hooyeeevan2511@gmail.com}  

\author{Ying Shan}
\affiliation{%
  \institution{ARC Lab, Tencent PCG}
  \state{Shenzhen}
  \country{China}}
\email{yingsshan@tencent.com}  

\author{Xiaohu Qie}
\affiliation{%
  \institution{Tencent PCG}
  \state{Shenzhen}
  \country{China}}
\email{tiger.qie@gmail.com} 

\author{Song-Hai Zhang}
\affiliation{%
  \institution{Tsinghua University}
  \state{Beijing}
  \country{China}}
\email{shz@tsinghua.edu.cn}  

\renewcommand{\shortauthors}{Guo et al.}

\begin{abstract} 
 With the emergence of neural radiance fields (NeRFs), view synthesis quality has reached an unprecedented level. Compared to traditional mesh-based assets, this volumetric representation is more powerful in expressing scene geometry but inevitably suffers from high rendering costs and can hardly be involved in further processes like editing, posing significant difficulties in combination with the existing graphics pipeline. In this paper, we present a hybrid volume-mesh representation, VMesh, which depicts an object with a textured mesh along with an auxiliary sparse volume. VMesh retains the advantages of mesh-based assets, such as efficient rendering, compact storage, and easy editing, while also incorporating the ability to represent subtle geometric structures provided by the volumetric counterpart. VMesh can be obtained from multi-view images of an object and renders at 2K 60FPS on common consumer devices with high fidelity, unleashing new opportunities for real-time immersive applications. Interactive demos are available on our project page: \textbf{\textcolor{teaser-green}{\url{https://bennyguo.github.io/vmesh/}}}.
\end{abstract}




\maketitle

\begin{figure}
    \centering
    \includegraphics[width=0.9\linewidth]{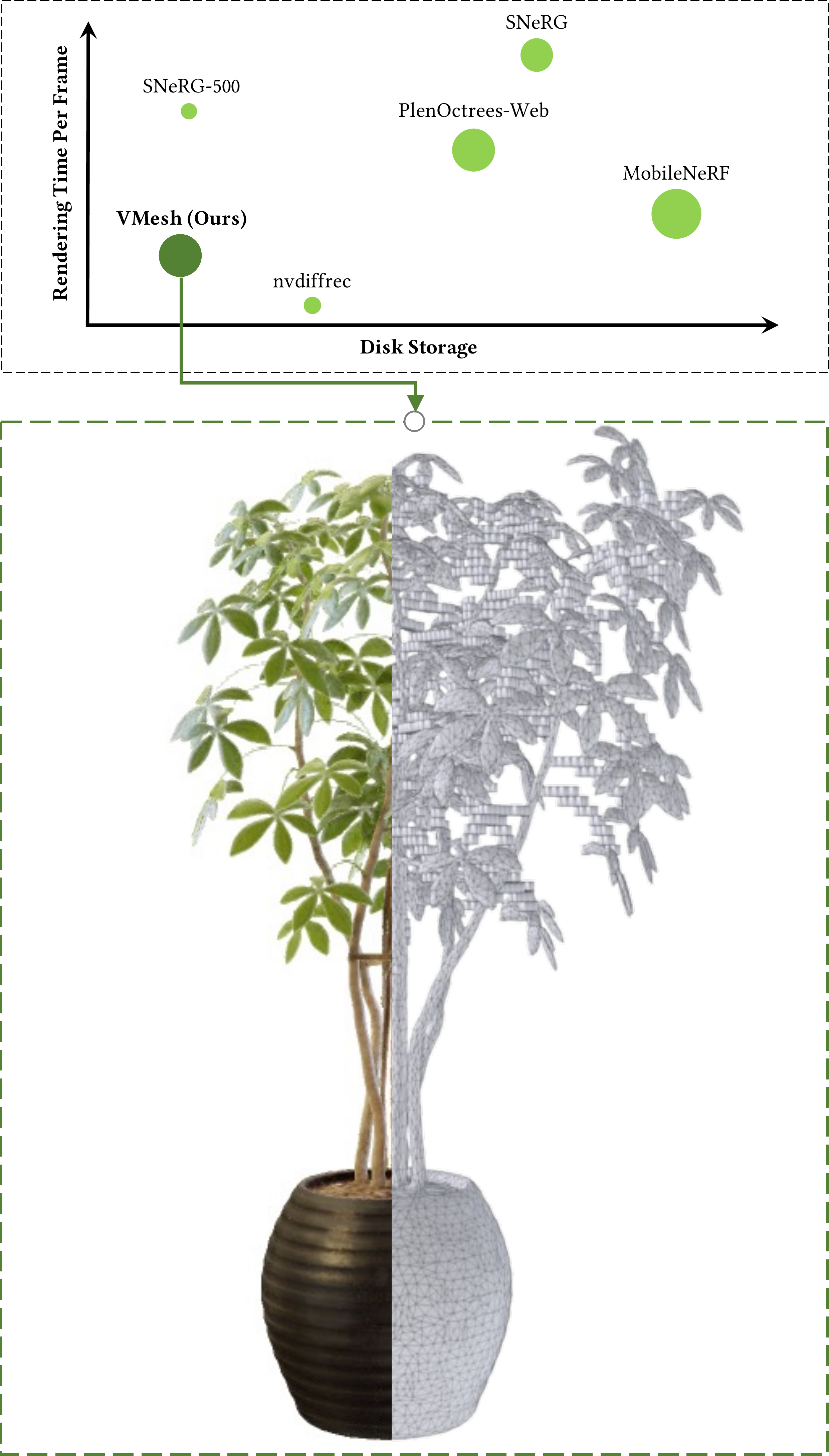}
    \caption{Our hybrid volume-mesh representation, VMesh, depicts object geometry with a triangular mesh and a sparse volume (cubes in the wireframe rendering). In the bubble chart, the area of the circle is proportional to the view synthesis quality on the NeRF-Synthetic dataset. VMesh achieves comparable view synthesis quality with existing real-time NeRF variants while being significantly more efficient in rendering speed and storage cost. }
    \label{fig:teaser}
\end{figure}

\section{Introduction}
Modern graphics engines mostly rely on polygonal meshes for efficient scene representation and rendering. However, when it comes to the problem of novel view synthesis, the rendering quality of mesh-based representations is greatly limited by the quality of the reconstructed geometry. On the other hand, volumetric representations like NeRF~\cite{nerf} have gained more and more attention for their superior view synthesis quality, but at the cost of slow rendering. Although researchers have devoted extensive efforts to enable real-time rendering of volumetric representations~\cite{fastnerf,plenoctrees,snerg,ngp,mobilenerf}, the storage of large amounts of volumetric data often causes severe memory issues, which prevents them from displaying on low-end devices or scaling to high-resolution scenes.

In this paper, we pursue a better representation of the geometry and appearance of an object for view synthesis from multi-view images. The representation should have the following characteristics:
\begin{itemize}
    \item Able to be rendered at a resolution of modern screens (1080P) in real-time (30FPS) on consumer-grade devices (laptops, tablets, mobile phones).
    \item Efficient in storage to model an object, occupying moderate disk space (<50MB).
    \item Has enough capability to model detailed geometry (such as thin structures) and view-dependent effects.
\end{itemize}
Besides these primary goals, we also expect the representation to be editable in some ways and have the potential to be integrated into modern graphics engines.

Mesh-based assets commonly adopted in graphics pipelines would meet most of these goals, i.e., high efficiency in rendering and storage, and great editability.
However, obtaining high-quality meshes and accurate textures of an object from casually-captured images is still an open problem.
On the other hand, the volumetric representation used by NeRF-based methods achieves remarkable performance for modeling object appearance, while being less competitive in all other aspects. Therefore, to achieve our goals, instead of improving the volumetric representation to have the abilities it initially does not have, it is more natural to base on the mesh representation to enhance its capability in geometry and appearance modeling with volumetric primitives. The key here is that the volumetric primitives are expected to be very sparse, modeling only the parts that the mesh struggles to model, e.g., thin structures, and only exist when having enough impact on image quality. In this way, the overhead brought by the volumetric part could be minimized, making it possible for the representation to achieve high efficiency and expressiveness at the same time. 

In practice, if we already have a mesh and a volume, combining the two to render jointly could simply be done by raymarching where the ray terminates at mesh surfaces, and compositing the volume color and surface color. However, the main difficulty lies in obtaining the mesh and volume from multi-view images, which is a non-trivial process, as direct optimization of either a mesh or an explicit volumetric grid can be very challenging under the multi-view setting~\cite{nvdiffrec,directvoxgo,plenoxels}. We thus draw inspiration from recent implicit neural surface reconstruction methods~\cite{volsdf,neus} where signed distance values can be converted to volume densities, and propose to obtain our hybrid representation from an implicit signed distance-density field. In this way, the signed distance field which represents a surface, and the density field which represents a volume can be jointly optimized by volume rendering. Through subsequent quality-preserving quantization steps, the signed distance field and the density field are converted to a triangular mesh and a sparse volume respectively. Furthermore, to model view-dependent effects while enabling efficient rendering and storage, we propose a RefBasis texture representation that mimics physically-based rendering. By representing the view-dependent specular color as the multiplication of a specular tint, a Fresnel attenuation factor, and an environment light given by learned basis functions, our texture representation achieves the best visual quality among existing real-time alternatives with significantly lower storage demand.

We implement a WebGL renderer for our VMesh representation and show that VMesh assets can be rendered in real-time frame rates at HD resolution on consumer-grade devices like mobile phones, tablets, and laptops. It achieves comparable view synthesis quality to existing real-time NeRF variants on various challenging scenes while offering very low storage costs. The explicit mesh surface also facilitates a wide range of applications, including shape deformation and texture editing.

\section{Related Work}

\subsection{Real-Time View Synthesis}
Real-time rendering has long been an important goal pursued by view synthesis techniques, especially the ones based on neural rendering. Since NeRF~\cite{nerf} emerged as the new quality standard for view synthesis, many attempts have been made to achieve efficient rendering with NeRF while preserving its high quality. Although many of them succeed in efficient rendering on high-end GPUs, few can maintain real-time performance on consumer-grade devices. PlenOctrees~\cite{plenoctrees} utilizes spherical harmonics to represent the view-dependent color, enabling storing volume density and color information in a sparse grid organized by octrees, therefore eliminating the massive computational cost of MLP evaluations. SNeRG~\cite{snerg} applies a deferred rendering scheme so that the view-dependent color only needs to be computed once for each pixel. It uses a very small MLP for color computation to balance the quality and rendering speed. To further improve rendering efficiency, MobileNeRF~\cite{mobilenerf} optimizes a set of triangle faces instead of a volumetric field, taking advantage of the high efficiency of the graphics pipeline. It adopts a similar texture representation as SNeRG and is applicable on common devices like laptops and mobile phones. However, the improvement in rendering efficiency introduces trade-offs with storage. These methods often require hundreds or even thousands of megabytes of disk storage for a single object, as storing densely sampled volumetric data or irregular triangles is not efficient. Moreover, they all lack a surface representation, which makes it hard to support further processing such as editing, simulation, shadow cast, and collision detection, to name a few. VMesh proposed in this paper is based on a textured mesh and only relies on volumetric primitives to enhance its representational ability in certain regions, which makes it both memory-efficient and processing-friendly. Our work is different from those mesh-based inverse rendering methods in that we do not assume a strict physically-based rendering model. Even so, we take inspiration from these methods to design our texture representation, which is demonstrated to be powerful enough for high-quality view synthesis while being editable.

\subsection{Hybrid Volume-Mesh Representation}
Mesh has been widely utilized as geometry guidance for NeRF variants to achieve geometry editing or handle deformable objects. NeRF-Editing~\cite{nerf-editing} allows controllable shape deformation on a NeRF representation by first deforming the extracted mesh and then training a deformation field based on the mesh deformation. NeuMesh~\cite{neumesh} learns vertex features for an extracted mesh and uses it for volume rendering, which also allows view synthesis of deformed objects by directly deforming the proxy mesh. Methods for building head avatars~\cite{instant-volumetric-head-avatars} often take advantage of mesh-based parametric face models to guide the ray marching process. EyeNeRF~\cite{eyenerf} uses an explicit eyeball surface to compute reflection and refraction rays for ray marching. Neural Assets~\cite{neural-assets} extracts a proxy mesh from a trained NeRF to support shadow cast and collision detection. These methods only make use of the geometry of the mesh and solely rely on volume rendering to produce the visual appearance. In contrast, our VMesh representation uses mesh and volume to represent separate geometries, and the mesh can be directly used for rendering. The work of Loubet \textit{et al.}~\cite{hybrid-mesh-volume-lods} is most related to ours, where a hybrid volume-mesh representation is used to build LoDs for complex 3D assets as the mesh is inefficient for pre-filtering sub-resolution structures, especially at large LoD scales. The authors introduce a heuristic approach to automatically find sub-resolution geometry in a mesh, and perform pre-filtering by voxelization. We share the same intuition to represent subtle structures with volume but manage to obtain such representation from multi-view images for view synthesis instead of processing existing assets.

\begin{figure*}[h]
  \centering
  \includegraphics[width=\linewidth]{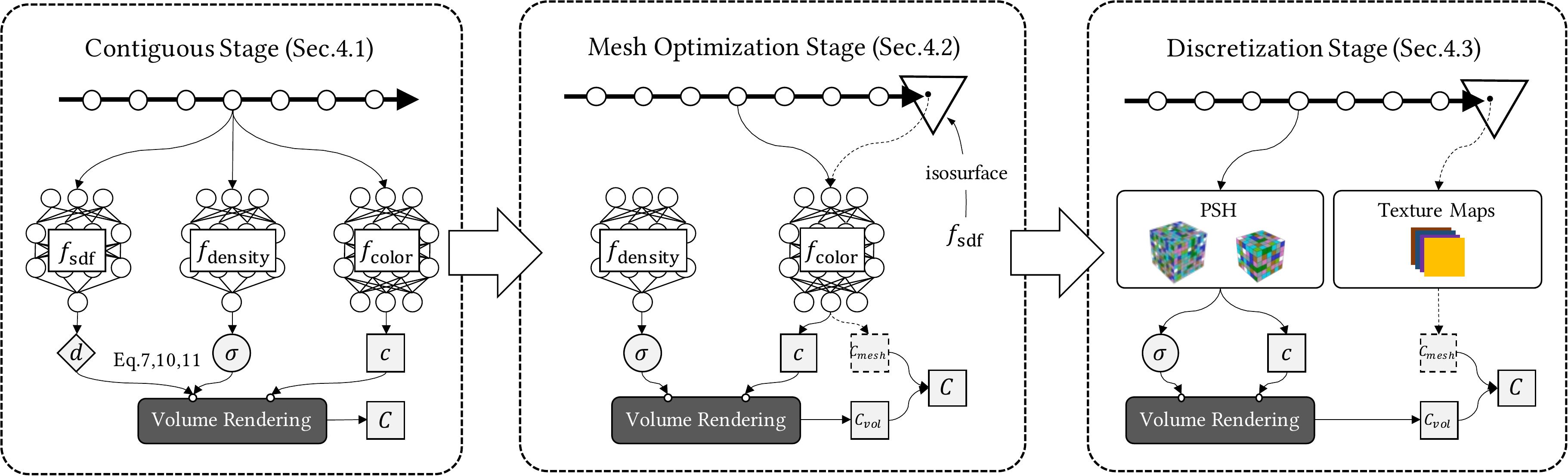}
  \caption{The 3-stage pipeline of training VMesh representation from multi-view images. We propose to start with an implicit signed distance-density field and gradually discretize the geometry and texture to obtain assets for real-time rendering.}
  \label{fig:pipeline}
\end{figure*}

\section{Preliminaries}

\subsection{Neural Radiance Fields}
Neural Radiance Fields (NeRF)~\cite{nerf} is the state-of-the-art method for novel view synthesis. It represents a scene as a continuous volumetric field modeled by a multi-layer perceptron (MLP). For each spatial location $\bm{x}=(x,y,z)$ and viewing direction $\bm{d}=(\theta,\phi)$, the networks output the volume density $\sigma$ and view-dependent color $\bm{c}$:
\begin{equation}
\begin{split}
    f_{\text{density}}&: \bm{x} \rightarrow \sigma\\
    f_{\text{color}}&: (\bm{x},\bm{d}) \rightarrow \bm{c}
\end{split}
\end{equation}
For each pixel to be rendered, the network is first evaluated on all sampled points $t_i$ along the camera ray $\bm{x}(t)=\bm{o}+t\bm{d}$ to get their densities and color. The pixel color $\hat{C}$ is estimated by the quadrature approximation of the volume rendering equation:
\begin{equation}\label{eqn:volume-rendering}
\begin{split}
\hat{\bm{C}}(\bm{r};\Theta)&=\sum_{i}\omega_i\bm{c}_i=\sum_{i}T_i\alpha_i\bm{c}_i \\
\alpha_i&=1-\text{exp}(-\sigma_i\delta_i) \\
T_i&=\prod_{j<i}(1-\alpha_j)
\end{split}
\end{equation}
where $\delta_i=t_{i+1}-t_i$. NeRF is optimized by minimizing the error between the rendered color $\hat{\bm{C}}$ and ground truth color $\bm{C}$:
\begin{equation} \label{eqn:photometric-loss}
\mathcal{L}=\sum_i||\hat{\bm{C}}_i-\bm{C}_i||_2
\end{equation}

\subsection{NeuS}
NeuS~\cite{neus} is a surface reconstruction method based on neural implicit representation. It represents the surface of an object as the zero-level set of a signed distance field (SDF) modeled by an MLP $f_{\Theta}$:
\begin{equation}
    f_{\text{sdf}}: \bm{x} \rightarrow d
\end{equation}
To effectively train the SDF, NeuS derives an unbiased mapping in the first order of approximation from the signed distance value $d$ to an opaque density $\rho$, the counterpart of the volume density $\sigma$:
\begin{equation} \label{eqn:opaque-density-neus}
    \rho(\bm{x}(t)) = \max \left(\frac{-\frac{\mathrm{d} \Phi_s}{\mathrm{~d} t}(f_{\text{sdf}}(\bm{x}(t)))}{\Phi_s(f_{\text{sdf}}(\bm{x}(t)))}, 0\right)
\end{equation}
where $\Phi_s$ is the Sigmoid function
\begin{equation} \label{eqn:sigmoid-neus}
\Phi_s(x)=\left(1+e^{-s x}\right)^{-1}
\end{equation}
with $s$ as a scene-wise learnable parameter. $1/s$ could be viewed as the uncertainty of the surface, with $1/s=0$ being equivalent to performing surface rendering. The same volume rendering equation (\cref{eqn:volume-rendering}) can be applied to get the pixel color with a discretized version of volume opacity $\alpha_i$ given by:
\begin{equation}
    \alpha_i=\max\left (\frac{\Phi_s(d_i)-\Phi_s(d_{i+1})}{\Phi_s(d_i)}, 0\right)
\end{equation}
Apart from the photometric loss in \cref{eqn:photometric-loss}, NeuS also utilizes an Eikonal term to regularize the SDF:
\begin{equation}
    \mathcal{L}_{\text{eik}}=\sum_{i}(||\nabla f_{\text{sdf}}(\bm{x}(t_i))||-1)^2
\end{equation}
and an optional object mask loss:
\begin{equation}
    \mathcal{L}_{\text{mask}}=\text{BCE}(\hat{M}, M)
\end{equation}
where $\hat{M}=\sum_{i}\omega_i$ is the estimated opacity along a ray, $M$ is the ground truth mask, and BCE is binary cross entropy.

\section{Hybrid Volume-Mesh (VMesh)}

Our hybrid volume-mesh representation, or VMesh, depicts object geometry with a triangular mesh surface and a sparse volume. The two types of geometric representation share the same texture formulation and can be jointly rendered by raymarching. We propose to obtain such a representation from multi-view images of an object. To start with, we train a contiguous form of the representation (\cref{sec:contiguous-stage}), where the surface part is modeled by a neural signed distance field, and the volume part is modeled by a neural density field. Following NeuS, we first convert the signed distance value to volume opacity and naturally combine the two parts by volume rendering.
Then we fix the learned signed distance field and extract a triangular mesh from it as a substitution to be rendered jointly with the neural density field (\cref{sec:mesh-optimization-stage}). We utilize differentiable isosurface and rasterization techniques to get high-quality meshes that align well with the implicit geometry. Lastly, we drop all the neural networks and perform discretization to get the final assets for efficient storage and rendering (\cref{sec:discretized-stage}). Concretely, the triangular mesh is simplified and UV-parametrized, and the neural density field is first voxelized and pruned to a sparse volume, which is then organized by perfect spatial hashing to support fast indexing and compact storage. The training pipeline is illustrated in \cref{fig:pipeline}. We further implement a WebGL2 renderer to enable real-time rendering in modern web browsers on various kinds of consumer-grade devices (\cref{sec:real-time-rendering}).

\subsection{Contiguous Stage} \label{sec:contiguous-stage}
We start from an implicit neural representation where each point in space holds two geometric properties, a signed distance value $d$ for the surface and a density value $\sigma$ for the volume. According to \cref{eqn:opaque-density-neus}, $d$ could be converted to volume opacity $\alpha$, allowing us to treat the surface part to be also a volume. To make the two parts work together, we could simply overlay the two volumes, summing up their densities:
\begin{equation}
    \sigma_{\text{hyb}} = \sigma_{\text{surf}} + \sigma_{\text{vol}}
\end{equation}
According to the definition of volume opacity $\alpha$, the above equation is equivalent to:
\begin{equation} \label{eqn:hybrid-opacity}
    \alpha_{\text{hyb}} = 1 - (1 - \alpha_{\text{surf}})(1 - \alpha_{\text{vol}})
\end{equation}
where $\alpha_{\text{surf}}$ and $\alpha_{\text{vol}}$ can be obtained from $d$ and $\sigma$ following \cref{eqn:opaque-density-neus} and \cref{eqn:volume-rendering} respectively. We use $\alpha_{\text{hyb}}$ to render the hybrid scene composed of both the surface and the volume and also use $\alpha_{\text{surf}}$ to render the surface only.

Implicit neural scene representations often rely on a large MLP to model object appearances. Given the feature of a spatial location and a viewing direction, the MLP predicts the view-dependent color, which can be time-consuming. Existing real-time view synthesis methods improve the efficiency of color queries mainly by
\begin{enumerate}
    \item replacing the MLP with fixed basis functions, such as spherical harmonics or spherical gaussians~\cite{plenoctrees,physg}.
    \item replacing the MLP with learned basis functions~\cite{fastnerf,nex}. The functions are also modeled by an MLP, but they can be discretized under acceptable memory consumption as they only depend on viewing directions.
    \item applying deferred shading and using a tiny MLP~\cite{snerg,mobilenerf}. In this way, the MLP is only evaluated per pixel instead of per sample point.
\end{enumerate}
These solutions all have their disadvantages. Fixed basis functions (1) and learned basis functions (2) often require numerous coefficients to be stored at each point to achieve enough representational ability. For example, PlenOctrees~\cite{plenoctrees} adopts spherical harmonics up to degree 3, resulting in 16 coefficients per color channel, 48 in total; FastNeRF~\cite{fastnerf} models 8 basis functions, which corresponds to 8 coefficients per color channel, 24 in total. Tiny MLP (3) struggles to model high-frequency view-dependent effects due to its limited capacity and is much more computationally extensive than the other two. In this work, we present a new texture representation named RefBasis for real-time view synthesis that is both representative and efficient in storage. 
We make the color depend on the reflected ray direction $\bm{\omega}_{r}$ instead of the incident ray direction, or viewing direction $\bm{\omega}_{o}$. This makes view-dependencies easier to model as demonstrated in Ref-NeRF~\cite{ref-nerf}. $\bm{\omega}_{r}$ can be computed as:
\begin{equation} \label{eqn:reflection-direction}
    \bm{\omega}_{r} = 2(\bm{\omega}_o\cdot\bm{n}) - \bm{\omega}_{o}
\end{equation}
where the normal direction $\bm{n}$ can be obtained as the gradient direction of the signed distance field by automatic differentiation:
\begin{equation}
    \bm{n} = \nabla f_{\text{sdf}}(\bm{x})
\end{equation}
We adopt a similar color formulation as Ref-NeRF, where the view-dependent color $\bm{c}$ is the composition of a diffuse color $\bm{c}_d$, and the multiplication of a specular tint $\bm{s}$ and a specular color $\bm{c}_s$. In Ref-NeRF, $\bm{c}_s$ is modeled by an MLP, taking $\bm{\omega}_{r}$ as input, as well as a per-point feature vector to bring additional degrees of freedom, and the incident angle to compensate for Fresnel effects. To enable real-time rendering, we instead model $\bm{c}_s$ as learned basis functions. Specifically, we predict a set of $N$ base specular colors $\{\bm{c}_{s}^1,\bm{c}_{s}^2,...,\bm{c}_{s}^{N}\}$ for each $\bm{\omega}_{r}$ and an N-dimensional weighting vector $(w_1,w_2,...,w_N)$ for each point. $\bm{c}_{s}$ is computed as the weighted summation of these base colors:
\begin{equation}
    \bm{c}_{s}(\bm{\omega}_r) = \sum_{i=1}^{N} w_i \bm{c}_{s}^{i}
\end{equation}
To explain Fresnel effects, we also predict an attenuation factor $\mathcal{A}$ from the incident angle $\theta$ and a per-point ``metallic" property $m$. The final color $c$ can be formulated as:
\begin{equation}
    \bm{c} = \gamma(\bm{c}_d + \bm{s} \odot \mathcal{A}(\theta,m)\bm{c}_s(\hat{\bm{\omega}}_r))
\end{equation}
where $\gamma$ is a fixed tone-mapping function and $\odot$ denotes element-wise multiplication. The base specular colors and the attenuation factor are modeled by MLPs and will be discretized as look-up tables in later stages.

In practice, we also predict a normal direction for each location and use this normal to compute $\bm{\omega}_{r}$. By forcing the predicted normal $\hat{\bm{n}}$ to be close to the analytic normal $\bm{n}$, $\hat{\bm{n}}$ acts as a smoothness prior at the beginning of training and a low-pass filter to improve the normal quality:
\begin{equation}
    \mathcal{L}_{\text{norm}} = 1 - \hat{\bm{n}}\cdot\bm{n}
\end{equation}

Since our goal is to represent most parts of the scene with surfaces, and only to model the ``hard" areas with volumetric matters, we apply the photometric loss simultaneously on the hybrid rendered color $\hat{\bm{C}}_i^{\text{hyb}}$ with $\alpha_{\text{hyb}}$ in \cref{eqn:hybrid-opacity} and the color $\hat{\bm{C}}_i^{\text{surf}}$ rendered only from the surface model with $\alpha_{\text{surf}}$:
\begin{equation} \label{eqn:photometric-loss-contiguous}
    \mathcal{L}_{\text{pm}}=\sum_i\left(||\hat{\bm{C}}_i^{\text{hyb}}-\bm{C}_i||_2+||\hat{\bm{C}}_i^{\text{surf}}-\bm{C}_i||_2\right)
\end{equation}
This prioritizes the usage of surfaces over volume wherever possible. We also penalize for volume densities as in ~\cite{plenoctrees} to further encourage volume sparsity:
\begin{equation} \label{eqn:volume-sparsity}
    \mathcal{L}_{\text{sp}}=\sum_{i}|1-\text{exp}(-\lambda \sigma_i)|
\end{equation}
where ${\sigma_i}$ are volume densities at sampled locations, and $\lambda$ is a hyperparameter.

We utilize an opaque loss which regularizes the rendered opacity of the surface model $\hat{M}^{\text{surf}}_i$ to be either 0 or 1:
\begin{equation}
    \mathcal{L}_{\text{opaque}} = \sum_i - \left(\hat{M}^{\text{surf}}_i log(\hat{M}^{\text{surf}}_i)+(1-\hat{M}^{\text{surf}}_i) log(1-\hat{M}^{\text{surf}}_i)\right)
\end{equation}
This encourages $s$ in \cref{eqn:sigmoid-neus} to be as large as possible, resulting in confident surfaces which benefit surface extraction in later stages.

We optimize the weighted summation of these loss terms in the contiguous stage:
\begin{equation}
\begin{split}
    \mathcal{L}_{\text{cont}} &= \lambda_{\text{pm}}\mathcal{L}_{\text{pm}} + \lambda_{\text{mask}}\mathcal{L}_{\text{mask}} + \lambda_{eik}\mathcal{L}_{\text{eik}} \\
    &+ \lambda_{\text{norm}}\mathcal{L}_{\text{norm}} + \lambda_{\text{sp}}\mathcal{L}_{\text{sp}} + \lambda_{\text{opaque}}\mathcal{L}_{\text{opaque}}
\end{split}
\end{equation}

\subsection{Mesh Optimization Stage} \label{sec:mesh-optimization-stage}
Once the contiguous representation is trained, we move to a mesh optimization stage where we extract a triangular mesh from the signed distance field and render it jointly with the neural density field for optimization. A simple way to achieve this is to use isosurface techniques such as Marching Cubes~\cite{marching-cubes} to extract the mesh and directly use it for further optimizations. However, due to the limited marching resolution, the extracted mesh cannot perfectly align with the implicit geometry, causing missing structures and biased surfaces. Redundant volume would appear to compensate for the inaccuracies, bringing unnecessary storage overhead. We solve this problem by optimizing the geometry of the extracted mesh with Deep Marching Tetrahedra~\cite{dmtet} and differentiable rasterization~\cite{pytorch3d,nvdiffrast}. Specifically, we fix $f_{\text{sdf}}$ from the contiguous stage and initialize a new signed distance field $f_{\text{sdf}}^{'}$ from $f_{\text{sdf}}$. We apply Deep Marching Tetrahedra on a dense grid with learnable grid vertex potisions, and render the extracted mesh to get its silhouette and depth map using a differentiable rasterizer~\cite{nvdiffrast}. The silhouette and depth map are compared with the opacity and depth value rendered from $f_{\text{sdf}}$ to make the mesh geometry closer to the implicit surface. The constraints can be formulated as:
\begin{equation}
    \mathcal{L}_{\text{mesh}}=\sum_i\left(||\hat{M}_i^{\text{surf}}-\hat{M}_i^{\text{mesh}}||_1 + ||\hat{D}_i^{\text{surf}}-\hat{D}_i^{\text{mesh}}||_1\right)
\end{equation}

To render the extracted mesh jointly with the neural density field, we apply raymarching that terminates at the mesh surface. For each pixel to be rendered, if it is occupied by the mesh surface, we only sample points in front of the surface to accumulate densities and colors. Then we alpha-blend the volume-rendered color $\hat{\bm{C}}_{\text{vol}}$ with the rasterized pixel color $\hat{\bm{C}}_{\text{mesh}}$ regarding the mesh surface as totally opaque ($\alpha_{\text{mesh}}=1$):
\begin{equation}
    \hat{\bm{C}}_{\text{hyb}} = \hat{\bm{C}}_{\text{vol}} + (1 - \hat{M}_{\text{vol}})\hat{\bm{C}}_{\text{mesh}}
\end{equation}
where $\hat{M}_{\text{vol}}$ is the transparency of the volume, as is $\sum_i\omega_i$ in \cref{eqn:volume-rendering}.
Note that till now we still rely on the MLPs to obtain the predicted normals and view-dependent colors. In this stage, we supervise on $\hat{\bm{C}}_{\text{hyb}}$ to optimize the appearance of this semi-contiguous representation:
\begin{equation}
    \mathcal{L}_{\text{pm}}=\sum_i||\hat{\bm{C}}_i^{\text{hyb}}-\bm{C}_i||_2
\end{equation}

The overall loss function is:
\begin{equation}
    \mathcal{L}_{\text{mo}} = \lambda_{\text{pm}}\mathcal{L}_{\text{pm}} + \lambda_{\text{mesh}}\mathcal{L}_{\text{mesh}} + \lambda_{\text{norm}}\mathcal{L}_{\text{norm}} + \lambda_{\text{sp}}\mathcal{L}_{\text{sp}}
\end{equation}
where $\mathcal{L}_{\text{norm}}$ and $\mathcal{L}_{\text{sp}}$ are the same as in the contiguous stage.

\subsection{Discretization Stage} \label{sec:discretized-stage}
In this stage, we convert all the neural representations to explicit assets for real-time rendering and efficient storage.
\subsubsection{Texture Space Discretization}
Earlier stages rely on MLPs to compute the specular color for each reflected ray direction and the corresponding attenuation factor. To enable fast color computation, we first densely sample all directions and convert the specular color MLP to a set of cube maps. We call these cube maps ``base environment maps" as they act like the environment map in image-based lighting techniques. For the attenuation factor MLP, we densely sample the $(\theta,m)$ space and convert it to a 2D look-up table.

\subsubsection{Mesh Discretization}
\paragraph{Simplification.} The mesh extracted from dense tetrahedron grids has an excess number of vertices and faces even in flat areas, which brings computation and storage overhead. To cope with this problem, we simplify the generated mesh by Quadric Edge Collapse Decimation~\cite{mesh-simplify}. Experiment results in \cref{tab:comparison-geometry} demonstrate that appropriate levels of mesh simplification do not undermine view synthesis quality but can, in fact, enhance it.
\paragraph{Parametrization.} We perform UV unwrapping to get texture coordinates for the mesh vertices, and sample densely in the UV space to retrieve the 2D normal map and texture maps.

\subsubsection{Volume Discretization}
\paragraph{Voxelization.} We densely evaluate the volume density for an $N^3$ voxel grid, and then track the max contribution of each voxel by shooting rays from all pixels of the training images. The contribution is calculated as the weight $\omega$ in \cref{eqn:volume-rendering}. Voxels with contributions lower than a threshold are pruned. For the remaining voxels, we uniformly sample 64 points inside each voxel and evaluate their average density, normal, and textures as an anti-aliased estimation of the properties of the voxel.
\paragraph{Hashing.} To enable efficient access to volume properties at arbitrary spatial locations while optimizing storage cost, existing works mainly adopt advanced data structures like octrees~\cite{plenoctrees}, or pack occupied voxels into smaller dense blocks~\cite{snerg}. Considering that the volume in our representation is very sparse (typically $<0.1\%$), we choose the latter approach and further utilize Perfect Spatial Hashing (PSH)~\cite{psh} for compact storage and efficient indexing. Denote the positions of occupied blocks as set $S$, with $|S|=N$, PSH constructs
\begin{itemize}
    \item a 3D hash table $H$ of size $m=\bar{m}^3 \approx N$ to store the volume data
    \item a 3D offset table $\Phi$ of size $r=\bar{r}^3 = \sigma N$ where $1/6\le\sigma<1$
\end{itemize}
so that there exists a perfect spatial hash function $h(\bm{p})$ as an injective map when $\bm{p} \in S$, mapping $\bm{p}$ to a unique slot in the hash table $H$. The hash function $h$ is defined as
\begin{equation} \label{eqn:perfect-spatial-hashing}
    h(\bm{p}) = \left((\bm{p}\bmod{\bar{m}}) + (\Phi[\bm{p}]\bmod{\bar{r}})\right) \bmod{\bar{m}}
\end{equation}
In this way, we can access volume data at any arbitrary location in constant $\mathcal{O}(1)$ time while only requiring the extra storage of a much smaller offset table other than the data itself.

\subsubsection{Fine-tuning}
To mitigate the quality loss brought by the quantization process, especially the texture seams caused by the UV-parametrization and the aliasing caused by voxelization (see \cref{fig:ablation} for visual examples), we directly fine-tune the extracted assets. However, the optimization of explicit geometry and textures could easily fall into local minima as they are no longer constrained by the inductive bias or local smoothness of the neural networks. Therefore, we fix the mesh geometry and use the rendered normal and texture images from the mesh optimization stage as pseudo ground truth references. In addition to the pixel-wise MSE loss, we also adopt a VGG perceptual loss~\cite{perceptual-loss} to optimize for image quality. To further reduce storage costs, it is a common practice to re-scale and quantize the values in the assets to 8-bit integers and store them as images. Since the quantization operation is not differentiable, existing methods mainly make it a post-processing step, which may bring quality loss. We instead propose to optimize the assets with a differentiable quantization module, which gives the quantized value in the forward pass and keeps the gradient untouched during the backward pass. In this way, we can ensure the rendering outputs are the same during optimization and in the real-time renderer. After fine-tuning, we store the mesh as a .obj file compressed using Draco, and all other assets as PNG files including the normal map, texture maps, volume data in the hash table, and the offset table. For unconstrained access to the volume data, we also store the occupancy of the $N^3$ grid as an occupancy image, where the occupancy status of a voxel corresponds to a single bit in the image.

\subsection{Real-Time Rendering} \label{sec:real-time-rendering}
We implement the rendering process of VMesh assets in WebGL2 using the three.js library~\cite{threejs}. The full process requires four render passes:
\begin{itemize}
    \item In the first and second passes, we rasterize the font and back face of the object bounding box to get the valid interval for raymarching. This could also be done by ray-box intersection.
    \item In the third pass, we rasterize the mesh to get the normal image, feature image, and depth image, and calculate the mesh rendering output based on the texture model.
    \item In the final pass, we first determine the start and termination points for raymarching. The start point is the front face position from the first pass. The termination point is the closer one of the back face position from the second pass and the surface position from the third pass. Then we uniformly sample points in this interval to get the volume rendering output. The volume-rendered color is alpha-composited with the mesh color the get the final rendering result.
\end{itemize}

\subsection{Implementation Details} \label{sec:implementation-details}
We implement the training process using the PyTorch~\cite{pytorch} framework.  In the contiguous and mesh optimization stage, we adopt the hash grid encoding and acceleration techniques proposed in Instant-NGP~\cite{ngp}. To stabilize training, we adopt a progressive training strategy to mask features from high-level hash grids in the early stage of training. We show in ~\cref{sec:exp-ablation} that this simple strategy could effectively alleviate shape-radiance ambiguity and improve geometry quality. We choose $N=4$ for the RefBasis texture. The contiguous stage is trained with a batch size of $8,192$ rays for $80,000$ iterations. We set $\lambda_{\text{pm}}=10$, $\lambda_{\text{mask}}=0.1$, $\lambda_{\text{eik}}=0.1$, $\lambda_{\text{norm}}=0.1$, linearly increase $\lambda_{\text{sp}}$ from 0.01 to 0.1, and only apply $\lambda_{\text{opaque}}=0.1$ after $50,000$ iterations. The mesh optimization stage is trained with a batch size of $8,192$ rays for $10,000$ iterations. The resolution of the dense grid for marching tetrahedra is set to 350. We set $\lambda_{\text{pm}}=10$, $\lambda_{\text{mesh}}=1$, $\lambda_{\text{norm}}=0.1$, and $\lambda_{\text{sp}}=0.1$ in this stage. In the discretization stage, we reduce the number of mesh faces to $1/4$, apply UV-parametrization using xatlas~\cite{xatlas} and render the normal and texture maps at $1024\times1024$. We store the base environment maps as cube maps of resolution $512\times512$, and create the look-up table at $256\times256$. We voxelize the volume part using a $512^3$ dense grid and implement a custom PyTorch C++ extension for the Perfect Spatial Hashing algorithm. It takes around 2 hours to optimize for a scene on a single RTX 3090.
\section{Experiments}
We mainly conduct experiments on the NeRF-Synthetic~\cite{nerf} dataset, which contains 8 challenging scenes with thin structures, highly reflective materials, and complex texture patterns. As we aim at real-time free-viewpoint rendering on consumer devices, we select the baseline methods as follows:
\begin{itemize}
    \item Real-time NeRF variants, including \textbf{PlenOctrees}~\cite{plenoctrees}, \textbf{SNeRG}~\cite{snerg} and \textbf{MobileNeRF}~\cite{mobilenerf}. These methods all work without the need for high-end GPUs. We also compare with a compressed version of PlenOctrees, denoted as \textbf{PlenOctrees-Web}, and two SNeRG variants with lower voxel grid resolution, denoted as \textbf{SNeRG-750} and \textbf{SNeRG-500} respectively.
    \item \textbf{nvdiffrec}~\cite{nvdiffrec}, that extracts the 3D mesh, PBR material and lighting from images. The extracted assets can be directly rendered in traditional graphics pipelines.
\end{itemize}
In \cref{sec:exp-comparisons} we first compare VMesh with these alternatives on image quality, rendering speed at different resolutions on various kinds of consumer devices, and disk storage cost. We compare with \textbf{nvdiffrec} and \textbf{MobileNeRF} on geometry quality as they also produce polygonal meshes. Then we perform thorough ablation studies in \cref{sec:exp-ablation} to show how different design choices could affect the image quality and efficiency. We show some qualitative results on real-captured scenes in \cref{sec:real-captured-results} and present several applications based on our representation in \cref{sec:exp-applications}.

\subsection{Comparisons} \label{sec:exp-comparisons}
\begin{figure*}[h]
  \centering
  \includegraphics[width=\linewidth]{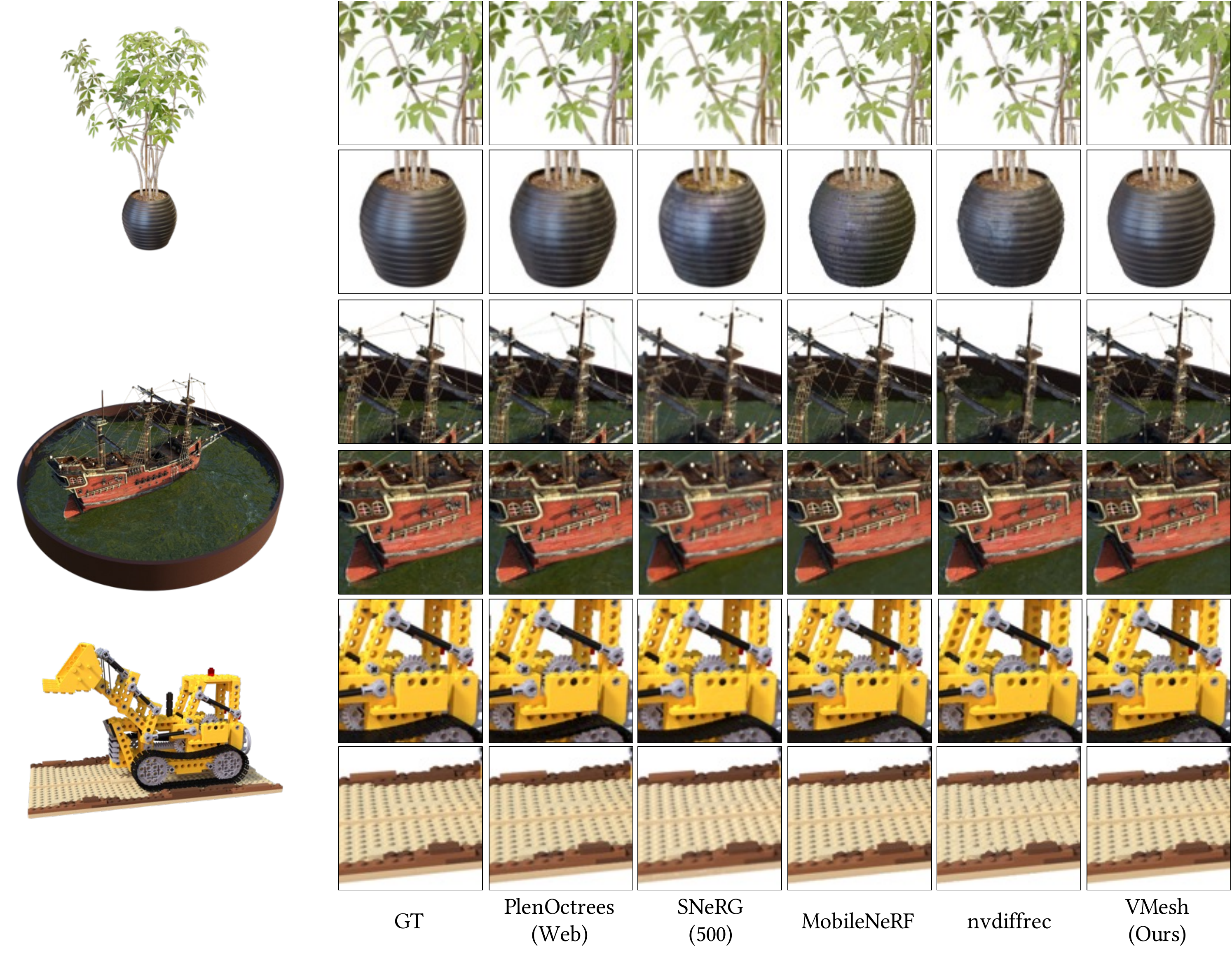}
  \caption{Qualitative comparison on NeRF-Synthetic test set.}
  \label{fig:comparison}
\end{figure*}
\begin{table}
\caption{Comparisons with state-of-the-art real-time view synthesis methods on NeRF-Synthetic dataset regarding image quality, rendering speed, and storage cost. The rendering speed is tested on a Macbook Pro (2020,M1) laptop. \\\footnotesize{*not supported on the tested device.}}
\label{tab:comparison}
\begin{tabular}{ c|ccccc } 
 Method & PSNR$\uparrow$ & SSIM$\uparrow$ & LPIPS$\downarrow$ & FPS$\uparrow$ & MB$\downarrow$ \\\hline
 NeRF & 31.00 & 0.947 & 0.081 & -* & 4.8 \\\hline
 PlenOctrees & 31.71 & 0.958 & 0.053 & -* & 1930 \\ 
 PlenOctrees-Web & 30.71 & 0.951 & 0.065 & 68 & 77.4 \\ 
 SNeRG & 30.38 & 0.950 & 0.050 & 43 & 86.8 \\ 
 SNeRG-750 & 29.94 & 0.947 & 0.064 & 44 & 44.7 \\ 
 SNeRG-500 & 28.93 & 0.939 & 0.064 & 56 & 17.8 \\ 
 MobileNeRF & 30.90 & 0.947 & 0.062 & 98 & 125.8 \\ 
 nvdiffrec & 29.05 & 0.939 & 0.081 & 616 & 41.9 \\\hline
 VMesh & 30.70 & 0.947 & 0.060 & 206 & 13.6 \\ 
\end{tabular}
\end{table}
\begin{table} 
\caption{Rendering efficiency comparison on different devices. VMesh is able to render at 2K 60FPS on a MacBook Pro (2020,M1). Note that the numbers are averaged across all scenes. Rendering framerates are capped at 60FPS on iPhone13 and iPad8. \\\footnotesize{*not supported on the tested device.}}
\label{tab:comparison-device}
\begin{tabular}{ c|c|c|c|c|c|c|c|c } 
 Device & \multicolumn{2}{c|}{Mobile} & \multicolumn{2}{|c|}{Tablet} & \multicolumn{4}{|c}{Laptop} \\\hline
 Model & \multicolumn{2}{c|}{iPhone13} & \multicolumn{2}{|c|}{iPad8} & \multicolumn{4}{|c}{MacBookPro (2020,M1)} \\\hline 
 Resolution & 800 & 1600 & 800 & 1600 & 800 & 1600 & 2048 & 4096 \\\hline
 SNeRG & -* & -* & -* & -* & 43 & 15 & 9 & 2 \\ 
 MobileNeRF & 57 & -* & 58 & 23 & 98 & 32 & 21 & 4 \\ 
 nvdiffrec & 60 & 60 & 60 & 60 & 616 & 342 & 226 & 63 \\\hline
 VMesh & 60 & 47 & 60 & 43 & 206 & 87 & 62 & 19 \\ 
\end{tabular}
\end{table}

We compare our VMesh representation with other alternatives on NeRF-Synthetic according to three aspects: (1) view synthesis quality on the test set in PSNR, SSIM and LPIPS; (2) rendering speed on a MacBook Pro (2020, M1) laptop measured with frames per second (FPS); (3) disk storage cost in megabytes. The results are shown in \cref{tab:comparison}, where the numbers are averaged across all 8 scenes. We also provide a bubble chart in \cref{fig:teaser} for more intuitive comparisons. In general, VMesh acts as a trade-off between volume rendering approaches (PlecOctrees, SNeRG, MobileNeRF) and mesh rendering approaches (nvdiffrec). Compared with volume rendering approaches, VMesh achieves competitive view synthesis quality with significantly higher rendering speed (2x faster than MobileNeRF, 5x faster than SNeRG) and lower storage cost (1/10 of MobileNeRF, 1/5 of SNeRG). The improvements could be attributed to the use of mesh for representing macro structures of the object, which is compact in storage and could fully utilize the efficiency of the graphics pipeline. Combined with our efficient texture representation, VMesh is able to render high-resolution images at real-time frame rates on various consumer-grade devices. In \cref{tab:comparison-device} we evaluate the rendering speed at different resolutions on three types of common devices: mobile phones, tablets, and laptops. nvdiffrec produces mesh assets with PBR materials, therefore serving as the upper bound for rendering efficiency. We can see from the table that the rendering speed of both SNeRG and MobileNeRF drops significantly as the resolution goes up, while VMesh maintains acceptable framerates even at 4K resolution. The reason for this discrepancy is that the computationally expensive raymarching processes or MLP evaluations are necessary for each pixel in SNeRG and MobileNeRF, which hinders efficient rendering at high resolutions. In contrast, our VMesh representation contains minimal amounts of volume, and our RefBasis texture representation is considerably more computationally efficient than tiny MLP. Some qualitative comparisons are shown in \cref{fig:comparison}, where we can see that VMesh does well in modeling view-dependent effects, and has enough representational ability in recovering subtle structures thanks to the volumetric counterpart.

\begin{figure}[h]
  \centering
  \includegraphics[width=\linewidth]{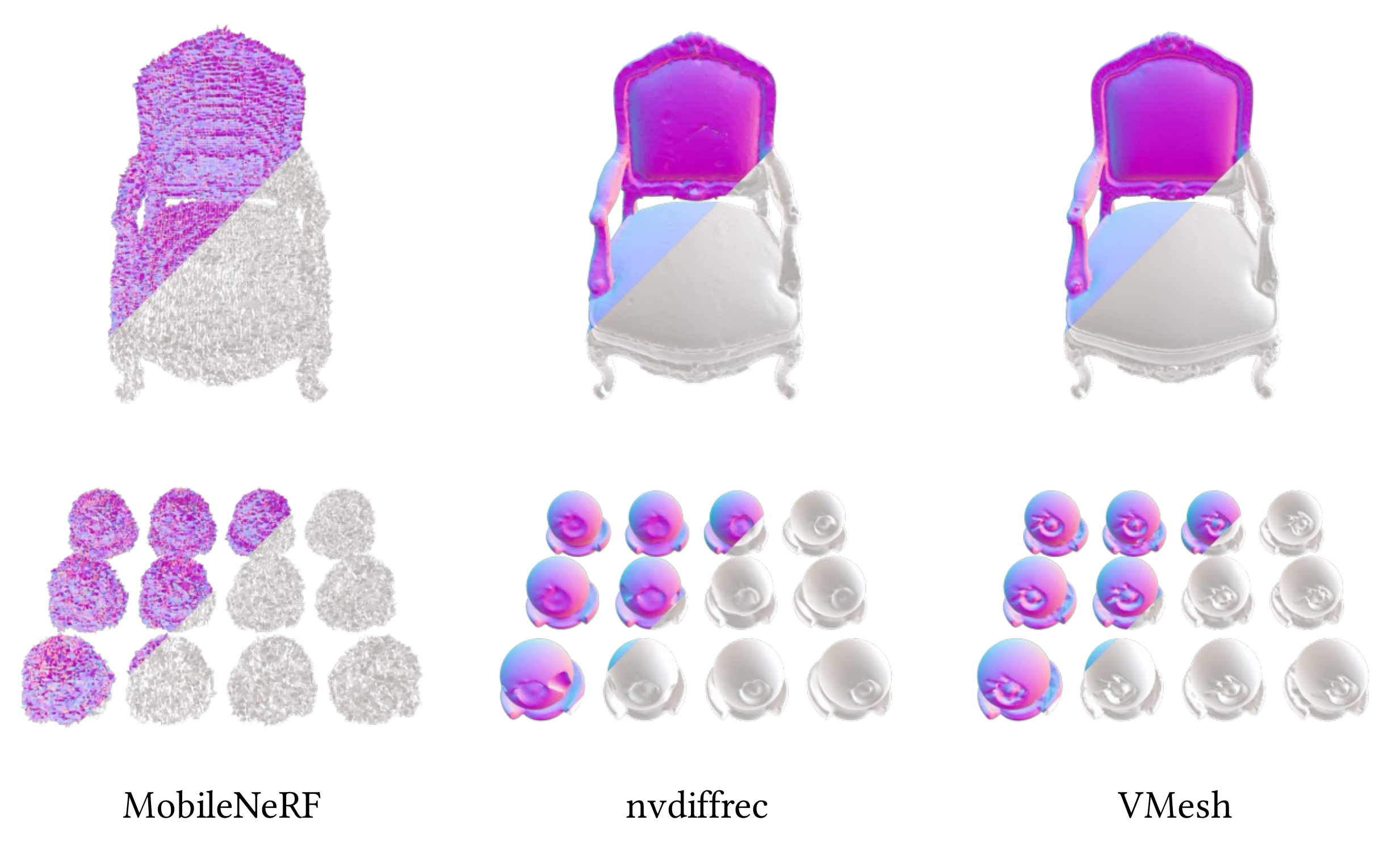}
  \caption{Qualitative comparison of the extracted mesh from MobileNeRF, nvdiffrec and VMesh. VMesh is able to extract meshes with high-quality surfaces.}
    \label{fig:comparison-geometry}
\end{figure}
\begin{table} 
\caption{Comparison on mesh complexity with MobileNeRF and nvdiffrec. The meshes extracted by VMesh can be drastically simplified without affecting visual quality.}
\label{tab:comparison-geometry}
\begin{tabular}{ c|c|ccc }
 \multicolumn{2}{c|}{Method} & PSNR$\uparrow$ & \#V$\downarrow$ & \#F$\downarrow$ \\\hline
 \multicolumn{2}{c|}{MobileNeRF} & 30.90 & 494,289 & 224,341 \\
 \multicolumn{2}{c|}{nvdiffrec} & 29.05 & 76,403 & 80,601 \\\hline
 \multirow{4}{*}{VMesh} & 1x & 30.48 & 385,464 & 771,275 \\
  & 2x & 30.66 & 192,637 & 385,620 \\
  & 4x & 30.70 & 96,231 & 192,809 \\
  & 8x & 30.62 & 48,028 & 96,404 \\
\end{tabular}
\end{table}
In \cref{fig:comparison-geometry}, we qualitatively compare the geometry quality with MobileNeRF and nvdiffrec as they also produce explicit triangular meshes. We also report the average number of vertices and faces on NeRF-Synthetic in \cref{tab:comparison-geometry}. It can be seen that our method achieves the best geometry quality among the three. MobileNeRF optimizes for separate triangles without any connection, which greatly limits its capability in tasks other than view synthesis. nvdiffrec produces meshes with unexpected holes. We believe this originates from the strict PBR assumption, which lacks the capability of explaining global illuminations, bringing geometry artifacts. VMesh creates high-quality meshes that can be drastically simplified without compromising rendering quality. This is due to the fact that we utilize meshes to represent larger, macro-level structures, which can be represented accurately using a smaller number of faces.

\subsection{Ablation Studies} \label{sec:exp-ablation}
\subsubsection{The necessity of volume}
\begin{figure}[h]
  \centering
  \includegraphics[width=\linewidth]{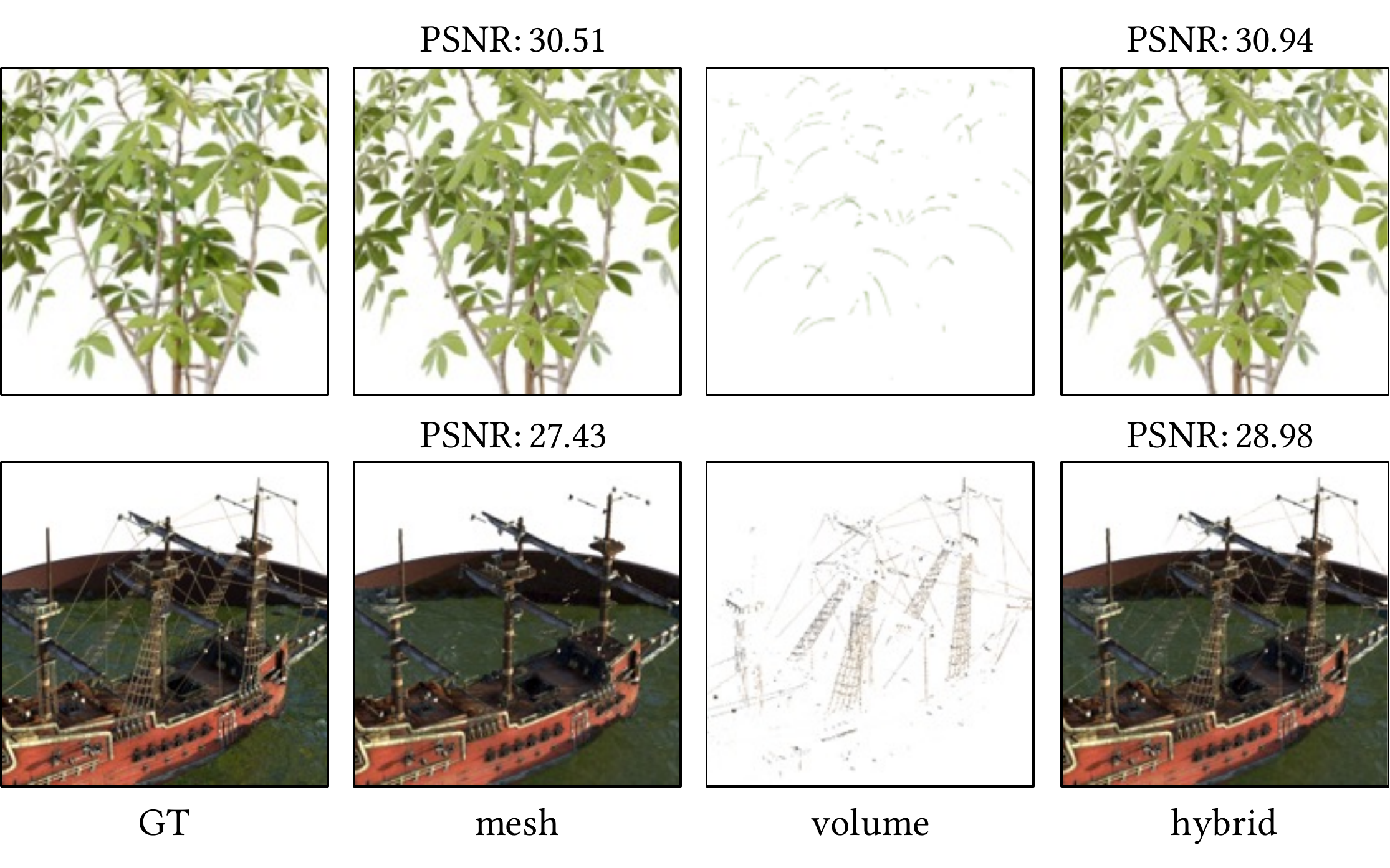}
  \caption{Qualitative explanation on the necessity of the volume part. Volumetric primitives could easily recover thin structures that are difficult to be modeled with mesh surfaces, like twigs and thin ropes.}
  \label{fig:ablation-hybrid}
\end{figure}
We first evaluate the necessity of the volumetric part of our VMesh representation. In \cref{fig:ablation-hybrid} we show results on the \textit{ficus} and \textit{ship} scene of the NeRF-Synthetic dataset. These two scenes contain objects with very subtle structures, such as twigs and thin ropes. With only the mesh part, these subtle structures are largely missing as they are hard (also inefficient) to be modeled with surfaces. Volumetric primitives, on the other hand, can be effective in handling these areas. Combining the two, VMesh can represent the challenging scenes very well.

\subsubsection{Texture representation}
\begin{table} 
\caption{Visual quality of different texture representations. We report the image quality metrics from the contiguous stage. $N$ denotes the number of features stored at each location. Q? denotes whether this representation can be quantized/cached for real-time rendering.}
\label{tab:ablation-texture}
\begin{tabular}{ c|ccccc } 
 Representation & PSNR$\uparrow$ & SSIM$\uparrow$ & LPIPS$\downarrow$ & N$\downarrow$ & Q? \\\hline
 NeRF & 31.53 & 0.959 & 0.057 & 16 & \xmark \\
 NeRF-R & 31.43 & 0.957 & 0.059 & 16 & \xmark \\
 IDR & 31.73 & 0.961 & 0.054 & 16 & \xmark \\ 
 Ref-NeRF & 31.50 & 0.957 & 0.057 & 11 & \xmark \\\hline
 SH (N=3) & 30.85 & 0.950 & 0.068 & 48 & \cmark \\ 
 SH-R (N=3) & 30.44 & 0.946 & 0.070 & 48 & \cmark \\ 
 LB (N=8) & 30.68 & 0.951 & 0.065 & 24 & \cmark \\ 
 LB-R (N=8) & 31.05 & 0.954 & 0.063 & 24 & \cmark \\\hline
 RefBasis (N=4) & 31.22 & 0.955 & 0.060 & 11 & \cmark \\ 
\end{tabular}
\end{table}
We evaluate different choices of texture representations in \cref{tab:ablation-texture} to demonstrate the effectiveness of our RefBasis Texture. We investigate five common types of texture representations:
\begin{itemize}
    \item NeRF~\cite{nerf} texture, which takes the geometry feature vector and the viewing direction as input, and models color by an MLP.
    \item IDR~\cite{idr} texture, which takes the geometry feature vector, the viewing direction, and the local normal direction as input, and models color by an MLP.
    \item Ref-NeRF~\cite{ref-nerf} texture, which takes the roughness value, the incident angle, the reflected ray direction, and the geometry feature vector as input, models the specular color by an MLP, and combines with the diffuse color and the specular tint to get the final color.
    \item Spherical Harmonics (SH) texture, which models color by coefficients of SH functions.
    \item Learned Basis (LB) texture, which models color by coefficients of a set of learned basis functions. The basis functions are modeled by an MLP, taking the viewing direction as input.
\end{itemize}
For a fair comparison, we also experiment with variants of these representations (marked with -R in the table) where the viewing direction is replaced with the reflected ray direction. We compare on view synthesis quality of the NeRF-Synthetic test set in the contiguous stage, as well as the number of features N stored at each location to indicate storage costs, and whether the representation can be quantized (Q? column in the table, check mark \cmark for quantizable) for real-time application. For representations that cannot be quantized, we choose N to be close to the one we use. And for SH and LB, we use commonly adopted settings in existing works. As shown in the table, among all quantizable representations, our RefBasis representation achieves the best visual quality with significantly lower storage costs. We achieve such efficiency by storing most of the common information in the neural environment map, which only has to be stored per scene instead of per point. In comparison to non-quantizable representations, RefBasis places greater emphasis on rendering efficiency at the expense of a minor decrease in visual quality.  

\subsubsection{Training strategies}
\begin{table} 
\caption{Ablation study on how design choices in different stages affect image quality.}
\label{tab:ablation}
\begin{tabular}{ c|c|ccc }
 Stage & Strategy & PSNR$\uparrow$ & SSIM$\uparrow$ & LPIPS$\downarrow$ \\\hline
 \multirow{4}{*}{\begin{tabular}{@{}c@{}}Contiguous\end{tabular}} & w/o pred. normal & 31.14 & 0.954 & 0.068 \\
  & w/o prog. training & 31.09 & 0.954 & 0.061 \\
  & w/o $\mathcal{L}_{\text{sp}}$ & 31.07 & 0.999 & 0.068 \\
  & Ours & 31.22 & 0.955 & 0.060 \\\hline
 \multirow{4}{*}{\begin{tabular}{@{}c@{}}Mesh\\Optimization\end{tabular}} & w/o mesh opt. & 29.81 & 0.947 & 0.071 \\
  & w/o learned grid vert. & 30.20 & 0.949 & 0.068 \\
  & w/o depth ref. & 30.72 & 0.952 & 0.063 \\
  & Ours & 30.89 & 0.953 & 0.062 \\\hline
 \multirow{2}{*}{\begin{tabular}{@{}c@{}}Discretization\end{tabular}} & w/o finetune & 28.50 & 0.919 & 0.100 \\
  & Ours & 30.70 & 0.947 & 0.060
\end{tabular}
\end{table}
\begin{figure*}[h]
  \centering
  \includegraphics[width=\linewidth]{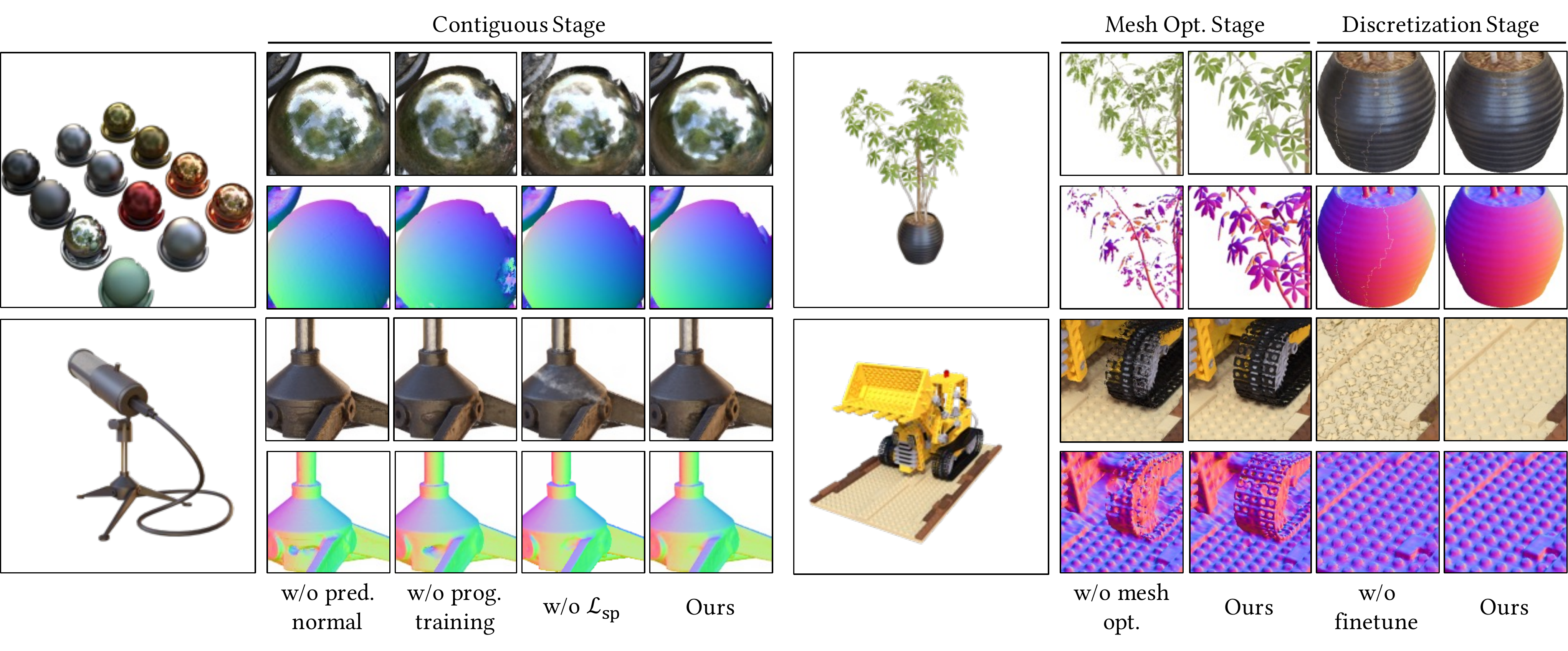}
  \caption{Qualitative visualization of the effect of different design choices.}
  \label{fig:ablation}
\end{figure*}
We conduct ablation studies on some of the training strategies to demonstrate their importance in achieving promising view synthesis quality, as shown in \cref{fig:ablation} and \cref{tab:ablation}. Without predicted normal (w/o pred. normal), the normal calculated by automatic differentiation can be noisy, leading to inaccurate reflection directions. Removing the volume sparsity regularization in \cref{eqn:volume-sparsity} (w/o $\mathcal{L}_\text{sp}$) will allow the model to use volume to explain view-dependent effects, resulting in foggy blobs as can be seen in the figure. This will also bring unnecessary additional storage costs for the redundant volume content. Without the progressive training strategy for hash encoding illustrated in \cref{sec:implementation-details} (w/o prog. training), shape-radiance ambiguity is more likely to happen, causing incorrect geometry. In the mesh optimization stage, we show that it is very crucial to optimize the extracted mesh for high-quality view synthesis. As can be seen in the figure, the extracted mesh could be far from ideal, with missing structures and inaccurate surfaces. After optimization using silhouette and depth constraints from the contiguous stage, the mesh quality is largely improved. Even so, nearly all the quality loss comes from this stage, further demonstrating the importance of having accurate geometries. In the discretization stage, the fine-tuning step helps remove the seams for better visual quality.

\subsubsection{Storage optimization}
\begin{table}
\caption{Ablation study on how different mesh texture resolutions affect image quality and storage.}
\label{tab:ablation-storage}
\begin{tabular}{ c|cc }
 Texture Resolution & PSNR$\uparrow$ & MB$\downarrow$ \\\hline
 $1024\times 1024$ & 30.70 & 13.6 \\
 $2048\times 2048$ & 30.80 & 32.7 \\
\end{tabular}
\end{table}
For the organization of volume data, we find that using a block size of 16 generally gives the best storage efficiency. We evaluate the impact of different mesh texture sizes on storage cost and show the results in \cref{tab:ablation-storage}. Using a texture resolution of 2048 consumes significantly larger storage (2.5x) but only brings marginal improvement in rendering quality (by +0.1db in PSNR). As a result, we employ a texture resolution of 1024 for all the experiments.

\subsection{Results on Real-Captured Scenes} \label{sec:real-captured-results}
\begin{figure*}[h]
  \centering
  \includegraphics[width=\linewidth]{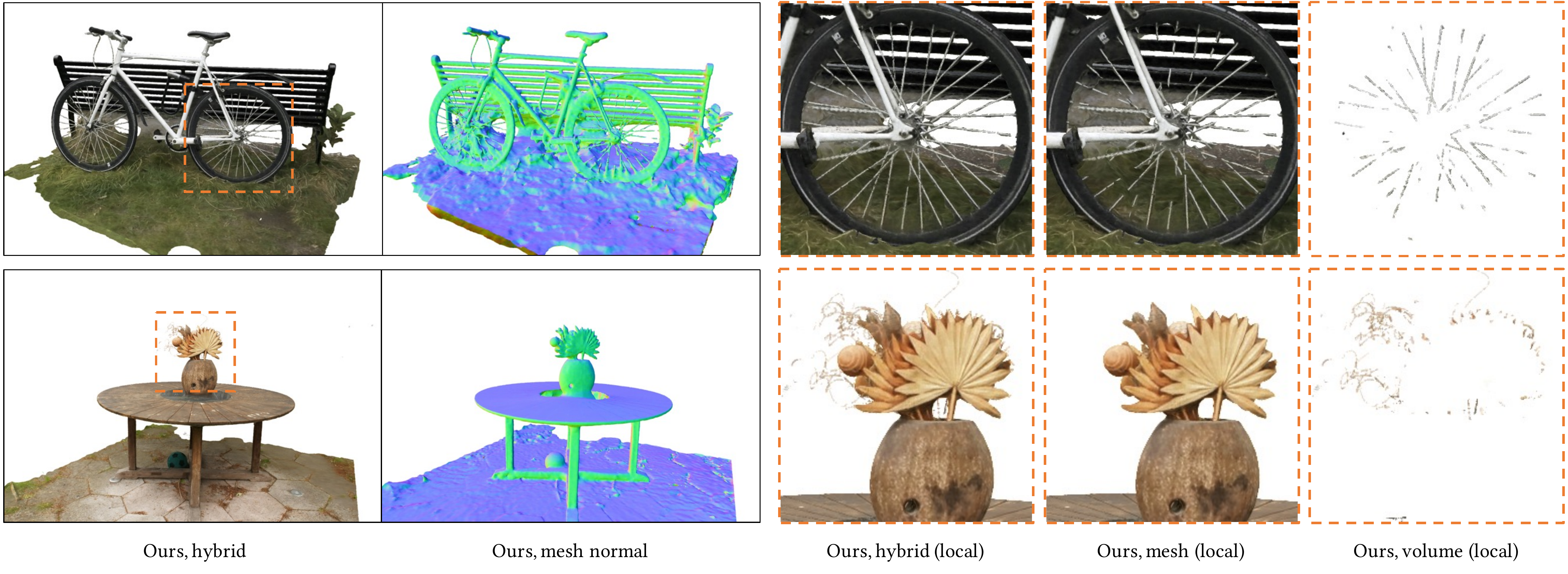}
  \caption{Qualitative results on two real-captured scenes. VMesh is able to model the thin structures of the foreground objects.}
  \label{fig:real-scenes}
\end{figure*}
In \cref{fig:real-scenes} we show qualitative results on real-captured scenes from Mip-NeRF 360~\cite{mipnerf360}. It is essential to point out that this paper only focuses on achieving efficient view synthesis of the foreground object. The background, however, is modeled using a similar approach to NeRF++~\cite{nerf++} and is excluded in the real-time renderer. VMesh successfully generalize on real-world scenes and is able to recover the subtle structures of the foreground objects such as the spokes on the bicycle wheels (up) and the fine filaments on dried flowers (down). 

\subsection{Applications} \label{sec:exp-applications}
\begin{figure}[h]
  \centering
  \includegraphics[width=\linewidth]{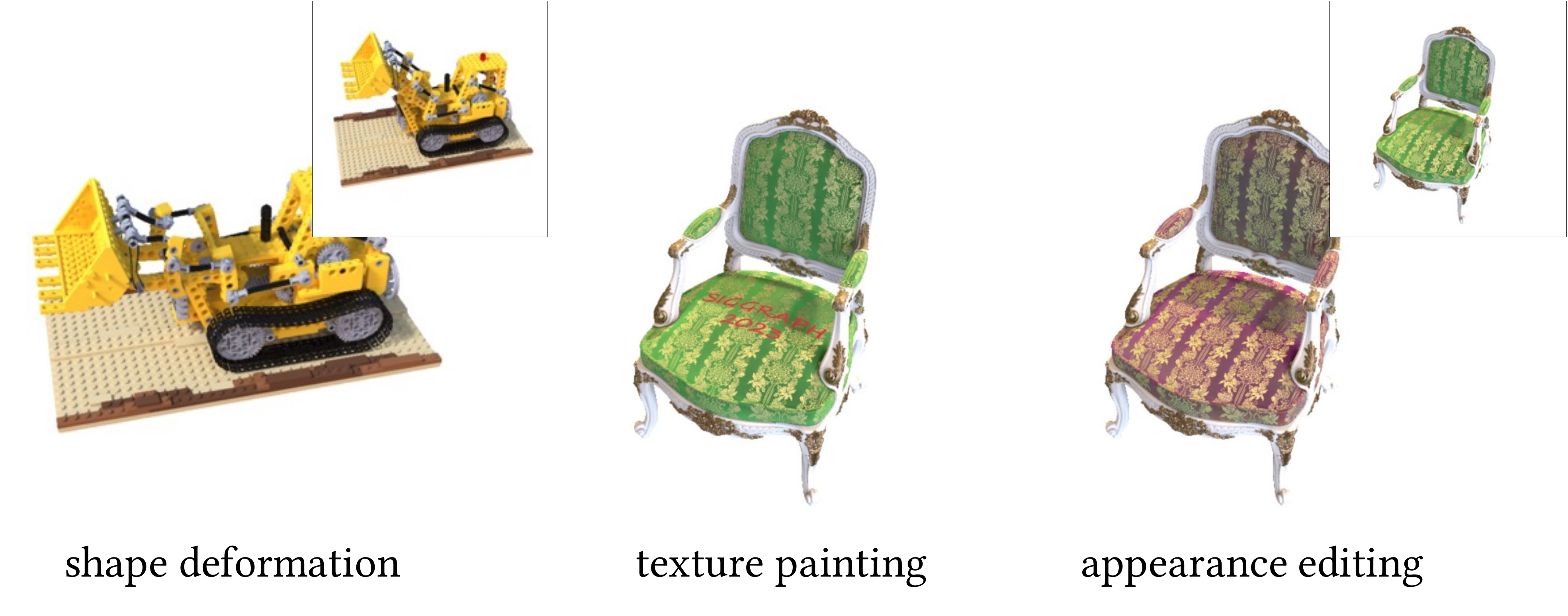}
  \caption{Applications on shape deformation, texture painting, and appearance editing. These tasks can all be easily accomplished thanks to the mesh geometry and the flexible RefBasis texture model.}
  \label{fig:applications}
\end{figure}
Benefiting from the mesh-based representation, we can intuitively perform various geometry and texture editing tasks which can be hard for volumetric-based methods. In \cref{fig:applications}  we show some examples of editing existing VMesh assets. In the shape deformation example (left), we use the ARAP (as-rigid-as-possible) method to deform the triangular mesh. Once the deformation is done, the deformed object can be instantly put into real-time view synthesis, without the need for re-training like in NeRF-Editing~\cite{nerf-editing}. In the texture painting (middle) and appearance editing (right) examples, we directly manipulate the diffuse and specular tint maps. This also shows that our RefBasis texture formulation is more flexible than existing alternatives like Spherical Harmonics or learned basis functions.

\section{Limitations}
There are several limitations inherent to the current VMesh representation that present opportunities for potential future improvements. These limitations include:
(1) The representation can produce inaccurate geometry due to global illumination effects caused by complex materials. For instance, in \cref{fig:real-scenes}, the tabletop displays unexpected holes as a result of the highly reflective, glass-like surface near its center. Our RefBasis texture formulation assumes approximate distant illumination and is therefore unable to accurately account for global illumination effects like self-reflections.
(2) VMesh is unable to precisely capture the appearance of transparent or furry objects. Ideally, these properties should be modeled by the volume part. However, we find that in practice, they are instead baked onto an inaccurate surface instead of being modeled by volume. This may result from using a surface rendering loss for optimization during the contiguous stage (see \cref{eqn:photometric-loss-contiguous}), and it requires a better surface-volume separation strategy to solve this problem.
(3) VMesh currently only facilitates real-time rendering of foreground objects, but there is potential to extend this hybrid formulation to efficiently model both foreground and background.

\section{Conclusions}
In summary, we present a new hybrid mesh-volume representation, VMesh, for efficient view synthesis of objects. VMesh combines the efficiency and flexibility of traditional mesh-based assets and the expressiveness of volumetric representations. We show competitive visual quality with state-of-the-art real-time view synthesis methods while being significantly faster and more efficient in storage. We believe VMesh may inspire future research into the potential of combining surface rendering and volume rendering techniques to produce high-quality view synthesis at a low cost.


\bibliographystyle{ACM-Reference-Format}
\bibliography{ref}


\begin{thebibliography}{33}


\ifx \showCODEN    \undefined \def \showCODEN     #1{\unskip}     \fi
\ifx \showDOI      \undefined \def \showDOI       #1{#1}\fi
\ifx \showISBNx    \undefined \def \showISBNx     #1{\unskip}     \fi
\ifx \showISBNxiii \undefined \def \showISBNxiii  #1{\unskip}     \fi
\ifx \showISSN     \undefined \def \showISSN      #1{\unskip}     \fi
\ifx \showLCCN     \undefined \def \showLCCN      #1{\unskip}     \fi
\ifx \shownote     \undefined \def \shownote      #1{#1}          \fi
\ifx \showarticletitle \undefined \def \showarticletitle #1{#1}   \fi
\ifx \showURL      \undefined \def \showURL       {\relax}        \fi
\providecommand\bibfield[2]{#2}
\providecommand\bibinfo[2]{#2}
\providecommand\natexlab[1]{#1}
\providecommand\showeprint[2][]{arXiv:#2}

\bibitem[Barron et~al\mbox{.}(2022)]%
        {mipnerf360}
\bibfield{author}{\bibinfo{person}{Jonathan~T. Barron}, \bibinfo{person}{Ben
  Mildenhall}, \bibinfo{person}{Dor Verbin}, \bibinfo{person}{Pratul~P.
  Srinivasan}, {and} \bibinfo{person}{Peter Hedman}.}
  \bibinfo{year}{2022}\natexlab{}.
\newblock \showarticletitle{Mip-NeRF 360: Unbounded Anti-Aliased Neural
  Radiance Fields}. In \bibinfo{booktitle}{\emph{{IEEE/CVF} Conference on
  Computer Vision and Pattern Recognition, {CVPR} 2022, New Orleans, LA, USA,
  June 18-24, 2022}}. \bibinfo{publisher}{{IEEE}}, \bibinfo{pages}{5460--5469}.
\newblock
\urldef\tempurl%
\url{https://doi.org/10.1109/CVPR52688.2022.00539}
\showDOI{\tempurl}


\bibitem[Bozic et~al\mbox{.}(2022)]%
        {neural-assets}
\bibfield{author}{\bibinfo{person}{Aljaz Bozic}, \bibinfo{person}{Denis
  Gladkov}, \bibinfo{person}{Luke Doukakis}, {and} \bibinfo{person}{Christoph
  Lassner}.} \bibinfo{year}{2022}\natexlab{}.
\newblock \showarticletitle{Neural Assets: Volumetric Object Capture and
  Rendering for Interactive Environments}.
\newblock \bibinfo{journal}{\emph{CoRR}}  \bibinfo{volume}{abs/2212.06125}
  (\bibinfo{year}{2022}).
\newblock
\urldef\tempurl%
\url{https://doi.org/10.48550/arXiv.2212.06125}
\showDOI{\tempurl}
\showeprint[arXiv]{2212.06125}


\bibitem[Chen et~al\mbox{.}(2022)]%
        {mobilenerf}
\bibfield{author}{\bibinfo{person}{Zhiqin Chen}, \bibinfo{person}{Thomas~A.
  Funkhouser}, \bibinfo{person}{Peter Hedman}, {and} \bibinfo{person}{Andrea
  Tagliasacchi}.} \bibinfo{year}{2022}\natexlab{}.
\newblock \showarticletitle{MobileNeRF: Exploiting the Polygon Rasterization
  Pipeline for Efficient Neural Field Rendering on Mobile Architectures}.
\newblock \bibinfo{journal}{\emph{CoRR}}  \bibinfo{volume}{abs/2208.00277}
  (\bibinfo{year}{2022}).
\newblock
\urldef\tempurl%
\url{https://doi.org/10.48550/arXiv.2208.00277}
\showDOI{\tempurl}
\showeprint[arXiv]{2208.00277}


\bibitem[Fridovich{-}Keil et~al\mbox{.}(2022)]%
        {plenoxels}
\bibfield{author}{\bibinfo{person}{Sara Fridovich{-}Keil},
  \bibinfo{person}{Alex Yu}, \bibinfo{person}{Matthew Tancik},
  \bibinfo{person}{Qinhong Chen}, \bibinfo{person}{Benjamin Recht}, {and}
  \bibinfo{person}{Angjoo Kanazawa}.} \bibinfo{year}{2022}\natexlab{}.
\newblock \showarticletitle{Plenoxels: Radiance Fields without Neural
  Networks}. In \bibinfo{booktitle}{\emph{{IEEE/CVF} Conference on Computer
  Vision and Pattern Recognition, {CVPR} 2022, New Orleans, LA, USA, June
  18-24, 2022}}. \bibinfo{publisher}{{IEEE}}, \bibinfo{pages}{5491--5500}.
\newblock
\urldef\tempurl%
\url{https://doi.org/10.1109/CVPR52688.2022.00542}
\showDOI{\tempurl}


\bibitem[Garbin et~al\mbox{.}(2021)]%
        {fastnerf}
\bibfield{author}{\bibinfo{person}{Stephan~J. Garbin}, \bibinfo{person}{Marek
  Kowalski}, \bibinfo{person}{Matthew Johnson}, \bibinfo{person}{Jamie
  Shotton}, {and} \bibinfo{person}{Julien P.~C. Valentin}.}
  \bibinfo{year}{2021}\natexlab{}.
\newblock \showarticletitle{FastNeRF: High-Fidelity Neural Rendering at
  200FPS}. In \bibinfo{booktitle}{\emph{{ICCV}}}. \bibinfo{publisher}{{IEEE}},
  \bibinfo{pages}{14326--14335}.
\newblock


\bibitem[Garland and Heckbert(1997)]%
        {mesh-simplify}
\bibfield{author}{\bibinfo{person}{Michael Garland} {and}
  \bibinfo{person}{Paul~S. Heckbert}.} \bibinfo{year}{1997}\natexlab{}.
\newblock \showarticletitle{Surface simplification using quadric error
  metrics}. In \bibinfo{booktitle}{\emph{Proceedings of the 24th Annual
  Conference on Computer Graphics and Interactive Techniques, {SIGGRAPH} 1997,
  Los Angeles, CA, USA, August 3-8, 1997}},
  \bibfield{editor}{\bibinfo{person}{G.~Scott Owen}, \bibinfo{person}{Turner
  Whitted}, {and} \bibinfo{person}{Barbara Mones{-}Hattal}} (Eds.).
  \bibinfo{publisher}{{ACM}}, \bibinfo{pages}{209--216}.
\newblock
\urldef\tempurl%
\url{https://doi.org/10.1145/258734.258849}
\showDOI{\tempurl}


\bibitem[Hedman et~al\mbox{.}(2021)]%
        {snerg}
\bibfield{author}{\bibinfo{person}{Peter Hedman}, \bibinfo{person}{Pratul~P.
  Srinivasan}, \bibinfo{person}{Ben Mildenhall}, \bibinfo{person}{Jonathan~T.
  Barron}, {and} \bibinfo{person}{Paul~E. Debevec}.}
  \bibinfo{year}{2021}\natexlab{}.
\newblock \showarticletitle{Baking Neural Radiance Fields for Real-Time View
  Synthesis}. In \bibinfo{booktitle}{\emph{{ICCV}}}.
  \bibinfo{publisher}{{IEEE}}, \bibinfo{pages}{5855--5864}.
\newblock


\bibitem[Johnson et~al\mbox{.}(2016)]%
        {perceptual-loss}
\bibfield{author}{\bibinfo{person}{Justin Johnson}, \bibinfo{person}{Alexandre
  Alahi}, {and} \bibinfo{person}{Li Fei{-}Fei}.}
  \bibinfo{year}{2016}\natexlab{}.
\newblock \showarticletitle{Perceptual Losses for Real-Time Style Transfer and
  Super-Resolution}. In \bibinfo{booktitle}{\emph{Computer Vision - {ECCV} 2016
  - 14th European Conference, Amsterdam, The Netherlands, October 11-14, 2016,
  Proceedings, Part {II}}} \emph{(\bibinfo{series}{Lecture Notes in Computer
  Science}, Vol.~\bibinfo{volume}{9906})},
  \bibfield{editor}{\bibinfo{person}{Bastian Leibe}, \bibinfo{person}{Jiri
  Matas}, \bibinfo{person}{Nicu Sebe}, {and} \bibinfo{person}{Max Welling}}
  (Eds.). \bibinfo{publisher}{Springer}, \bibinfo{pages}{694--711}.
\newblock
\urldef\tempurl%
\url{https://doi.org/10.1007/978-3-319-46475-6\_43}
\showDOI{\tempurl}


\bibitem[jpcy(2022)]%
        {xatlas}
\bibfield{author}{\bibinfo{person}{jpcy}.} \bibinfo{year}{2022}\natexlab{}.
\newblock \bibinfo{title}{xatlas}.
\newblock \bibinfo{howpublished}{\url{https://github.com/jpcy/xatlas}}.
\newblock


\bibitem[Laine et~al\mbox{.}(2020)]%
        {nvdiffrast}
\bibfield{author}{\bibinfo{person}{Samuli Laine}, \bibinfo{person}{Janne
  Hellsten}, \bibinfo{person}{Tero Karras}, \bibinfo{person}{Yeongho Seol},
  \bibinfo{person}{Jaakko Lehtinen}, {and} \bibinfo{person}{Timo Aila}.}
  \bibinfo{year}{2020}\natexlab{}.
\newblock \showarticletitle{Modular primitives for high-performance
  differentiable rendering}.
\newblock \bibinfo{journal}{\emph{{ACM} Trans. Graph.}} \bibinfo{volume}{39},
  \bibinfo{number}{6} (\bibinfo{year}{2020}), \bibinfo{pages}{194:1--194:14}.
\newblock
\urldef\tempurl%
\url{https://doi.org/10.1145/3414685.3417861}
\showDOI{\tempurl}


\bibitem[Lefebvre and Hoppe(2006)]%
        {psh}
\bibfield{author}{\bibinfo{person}{Sylvain Lefebvre} {and}
  \bibinfo{person}{Hugues Hoppe}.} \bibinfo{year}{2006}\natexlab{}.
\newblock \showarticletitle{Perfect spatial hashing}.
\newblock \bibinfo{journal}{\emph{{ACM} Trans. Graph.}} \bibinfo{volume}{25},
  \bibinfo{number}{3} (\bibinfo{year}{2006}), \bibinfo{pages}{579--588}.
\newblock
\urldef\tempurl%
\url{https://doi.org/10.1145/1141911.1141926}
\showDOI{\tempurl}


\bibitem[Li et~al\mbox{.}(2022)]%
        {eyenerf}
\bibfield{author}{\bibinfo{person}{Gengyan Li}, \bibinfo{person}{Abhimitra
  Meka}, \bibinfo{person}{Franziska M{\"{u}}ller}, \bibinfo{person}{Marcel~C.
  B{\"{u}}hler}, {and} \bibinfo{person}{Otmar Hilliges}.}
  \bibinfo{year}{2022}\natexlab{}.
\newblock \showarticletitle{EyeNeRF: {A} Hybrid Representation for
  Photorealistic Synthesis, Animation and Relighting of Human Eyes}.
\newblock \bibinfo{journal}{\emph{CoRR}}  \bibinfo{volume}{abs/2206.08428}
  (\bibinfo{year}{2022}).
\newblock


\bibitem[Lorensen and Cline(1987)]%
        {marching-cubes}
\bibfield{author}{\bibinfo{person}{William~E. Lorensen} {and}
  \bibinfo{person}{Harvey~E. Cline}.} \bibinfo{year}{1987}\natexlab{}.
\newblock \showarticletitle{Marching cubes: {A} high resolution 3D surface
  construction algorithm}. In \bibinfo{booktitle}{\emph{Proceedings of the 14th
  Annual Conference on Computer Graphics and Interactive Techniques, {SIGGRAPH}
  1987, Anaheim, California, USA, July 27-31, 1987}},
  \bibfield{editor}{\bibinfo{person}{Maureen~C. Stone}} (Ed.).
  \bibinfo{publisher}{{ACM}}, \bibinfo{pages}{163--169}.
\newblock
\urldef\tempurl%
\url{https://doi.org/10.1145/37401.37422}
\showDOI{\tempurl}


\bibitem[Loubet and Neyret(2017)]%
        {hybrid-mesh-volume-lods}
\bibfield{author}{\bibinfo{person}{Guillaume Loubet} {and}
  \bibinfo{person}{Fabrice Neyret}.} \bibinfo{year}{2017}\natexlab{}.
\newblock \showarticletitle{Hybrid mesh-volume LoDs for all-scale pre-filtering
  of complex 3D assets}.
\newblock \bibinfo{journal}{\emph{Comput. Graph. Forum}} \bibinfo{volume}{36},
  \bibinfo{number}{2} (\bibinfo{year}{2017}), \bibinfo{pages}{431--442}.
\newblock


\bibitem[Mildenhall et~al\mbox{.}(2022)]%
        {nerf}
\bibfield{author}{\bibinfo{person}{Ben Mildenhall}, \bibinfo{person}{Pratul~P.
  Srinivasan}, \bibinfo{person}{Matthew Tancik}, \bibinfo{person}{Jonathan~T.
  Barron}, \bibinfo{person}{Ravi Ramamoorthi}, {and} \bibinfo{person}{Ren Ng}.}
  \bibinfo{year}{2022}\natexlab{}.
\newblock \showarticletitle{NeRF: representing scenes as neural radiance fields
  for view synthesis}.
\newblock \bibinfo{journal}{\emph{Commun. {ACM}}} \bibinfo{volume}{65},
  \bibinfo{number}{1} (\bibinfo{year}{2022}), \bibinfo{pages}{99--106}.
\newblock


\bibitem[mrdoob(2023)]%
        {threejs}
\bibfield{author}{\bibinfo{person}{mrdoob}.} \bibinfo{year}{2023}\natexlab{}.
\newblock \bibinfo{title}{threejs}.
\newblock \bibinfo{howpublished}{\url{https://github.com/mrdoob/three.js}}.
\newblock


\bibitem[M{\"{u}}ller et~al\mbox{.}(2022)]%
        {ngp}
\bibfield{author}{\bibinfo{person}{Thomas M{\"{u}}ller}, \bibinfo{person}{Alex
  Evans}, \bibinfo{person}{Christoph Schied}, {and} \bibinfo{person}{Alexander
  Keller}.} \bibinfo{year}{2022}\natexlab{}.
\newblock \showarticletitle{Instant neural graphics primitives with a
  multiresolution hash encoding}.
\newblock \bibinfo{journal}{\emph{{ACM} Trans. Graph.}} \bibinfo{volume}{41},
  \bibinfo{number}{4} (\bibinfo{year}{2022}), \bibinfo{pages}{102:1--102:15}.
\newblock
\urldef\tempurl%
\url{https://doi.org/10.1145/3528223.3530127}
\showDOI{\tempurl}


\bibitem[Munkberg et~al\mbox{.}(2022)]%
        {nvdiffrec}
\bibfield{author}{\bibinfo{person}{Jacob Munkberg}, \bibinfo{person}{Wenzheng
  Chen}, \bibinfo{person}{Jon Hasselgren}, \bibinfo{person}{Alex Evans},
  \bibinfo{person}{Tianchang Shen}, \bibinfo{person}{Thomas M{\"{u}}ller},
  \bibinfo{person}{Jun Gao}, {and} \bibinfo{person}{Sanja Fidler}.}
  \bibinfo{year}{2022}\natexlab{}.
\newblock \showarticletitle{Extracting Triangular 3D Models, Materials, and
  Lighting From Images}. In \bibinfo{booktitle}{\emph{{IEEE/CVF} Conference on
  Computer Vision and Pattern Recognition, {CVPR} 2022, New Orleans, LA, USA,
  June 18-24, 2022}}. \bibinfo{publisher}{{IEEE}}, \bibinfo{pages}{8270--8280}.
\newblock
\urldef\tempurl%
\url{https://doi.org/10.1109/CVPR52688.2022.00810}
\showDOI{\tempurl}


\bibitem[Paszke et~al\mbox{.}(2019)]%
        {pytorch}
\bibfield{author}{\bibinfo{person}{Adam Paszke}, \bibinfo{person}{Sam Gross},
  \bibinfo{person}{Francisco Massa}, \bibinfo{person}{Adam Lerer},
  \bibinfo{person}{James Bradbury}, \bibinfo{person}{Gregory Chanan},
  \bibinfo{person}{Trevor Killeen}, \bibinfo{person}{Zeming Lin},
  \bibinfo{person}{Natalia Gimelshein}, \bibinfo{person}{Luca Antiga},
  \bibinfo{person}{Alban Desmaison}, \bibinfo{person}{Andreas K{\"{o}}pf},
  \bibinfo{person}{Edward~Z. Yang}, \bibinfo{person}{Zachary DeVito},
  \bibinfo{person}{Martin Raison}, \bibinfo{person}{Alykhan Tejani},
  \bibinfo{person}{Sasank Chilamkurthy}, \bibinfo{person}{Benoit Steiner},
  \bibinfo{person}{Lu Fang}, \bibinfo{person}{Junjie Bai}, {and}
  \bibinfo{person}{Soumith Chintala}.} \bibinfo{year}{2019}\natexlab{}.
\newblock \showarticletitle{PyTorch: An Imperative Style, High-Performance Deep
  Learning Library}. In \bibinfo{booktitle}{\emph{Advances in Neural
  Information Processing Systems 32: Annual Conference on Neural Information
  Processing Systems 2019, NeurIPS 2019, December 8-14, 2019, Vancouver, BC,
  Canada}}, \bibfield{editor}{\bibinfo{person}{Hanna~M. Wallach},
  \bibinfo{person}{Hugo Larochelle}, \bibinfo{person}{Alina Beygelzimer},
  \bibinfo{person}{Florence d'Alch{\'{e}}{-}Buc}, \bibinfo{person}{Emily~B.
  Fox}, {and} \bibinfo{person}{Roman Garnett}} (Eds.).
  \bibinfo{pages}{8024--8035}.
\newblock
\urldef\tempurl%
\url{https://proceedings.neurips.cc/paper/2019/hash/bdbca288fee7f92f2bfa9f7012727740-Abstract.html}
\showURL{%
\tempurl}


\bibitem[Ravi et~al\mbox{.}(2020)]%
        {pytorch3d}
\bibfield{author}{\bibinfo{person}{Nikhila Ravi}, \bibinfo{person}{Jeremy
  Reizenstein}, \bibinfo{person}{David Novotn{\'{y}}}, \bibinfo{person}{Taylor
  Gordon}, \bibinfo{person}{Wan{-}Yen Lo}, \bibinfo{person}{Justin Johnson},
  {and} \bibinfo{person}{Georgia Gkioxari}.} \bibinfo{year}{2020}\natexlab{}.
\newblock \showarticletitle{Accelerating 3D Deep Learning with PyTorch3D}.
\newblock \bibinfo{journal}{\emph{CoRR}}  \bibinfo{volume}{abs/2007.08501}
  (\bibinfo{year}{2020}).
\newblock
\showeprint[arXiv]{2007.08501}
\urldef\tempurl%
\url{https://arxiv.org/abs/2007.08501}
\showURL{%
\tempurl}


\bibitem[Shen et~al\mbox{.}(2021)]%
        {dmtet}
\bibfield{author}{\bibinfo{person}{Tianchang Shen}, \bibinfo{person}{Jun Gao},
  \bibinfo{person}{Kangxue Yin}, \bibinfo{person}{Ming{-}Yu Liu}, {and}
  \bibinfo{person}{Sanja Fidler}.} \bibinfo{year}{2021}\natexlab{}.
\newblock \showarticletitle{Deep Marching Tetrahedra: a Hybrid Representation
  for High-Resolution 3D Shape Synthesis}. In
  \bibinfo{booktitle}{\emph{Advances in Neural Information Processing Systems
  34: Annual Conference on Neural Information Processing Systems 2021, NeurIPS
  2021, December 6-14, 2021, virtual}},
  \bibfield{editor}{\bibinfo{person}{Marc'Aurelio Ranzato},
  \bibinfo{person}{Alina Beygelzimer}, \bibinfo{person}{Yann~N. Dauphin},
  \bibinfo{person}{Percy Liang}, {and} \bibinfo{person}{Jennifer~Wortman
  Vaughan}} (Eds.). \bibinfo{pages}{6087--6101}.
\newblock
\urldef\tempurl%
\url{https://proceedings.neurips.cc/paper/2021/hash/30a237d18c50f563cba4531f1db44acf-Abstract.html}
\showURL{%
\tempurl}


\bibitem[Sun et~al\mbox{.}(2022)]%
        {directvoxgo}
\bibfield{author}{\bibinfo{person}{Cheng Sun}, \bibinfo{person}{Min Sun}, {and}
  \bibinfo{person}{Hwann{-}Tzong Chen}.} \bibinfo{year}{2022}\natexlab{}.
\newblock \showarticletitle{Direct Voxel Grid Optimization: Super-fast
  Convergence for Radiance Fields Reconstruction}. In
  \bibinfo{booktitle}{\emph{{IEEE/CVF} Conference on Computer Vision and
  Pattern Recognition, {CVPR} 2022, New Orleans, LA, USA, June 18-24, 2022}}.
  \bibinfo{publisher}{{IEEE}}, \bibinfo{pages}{5449--5459}.
\newblock
\urldef\tempurl%
\url{https://doi.org/10.1109/CVPR52688.2022.00538}
\showDOI{\tempurl}


\bibitem[Verbin et~al\mbox{.}(2022)]%
        {ref-nerf}
\bibfield{author}{\bibinfo{person}{Dor Verbin}, \bibinfo{person}{Peter Hedman},
  \bibinfo{person}{Ben Mildenhall}, \bibinfo{person}{Todd~E. Zickler},
  \bibinfo{person}{Jonathan~T. Barron}, {and} \bibinfo{person}{Pratul~P.
  Srinivasan}.} \bibinfo{year}{2022}\natexlab{}.
\newblock \showarticletitle{Ref-NeRF: Structured View-Dependent Appearance for
  Neural Radiance Fields}. In \bibinfo{booktitle}{\emph{{IEEE/CVF} Conference
  on Computer Vision and Pattern Recognition, {CVPR} 2022, New Orleans, LA,
  USA, June 18-24, 2022}}. \bibinfo{publisher}{{IEEE}},
  \bibinfo{pages}{5481--5490}.
\newblock
\urldef\tempurl%
\url{https://doi.org/10.1109/CVPR52688.2022.00541}
\showDOI{\tempurl}


\bibitem[Wang et~al\mbox{.}(2021)]%
        {neus}
\bibfield{author}{\bibinfo{person}{Peng Wang}, \bibinfo{person}{Lingjie Liu},
  \bibinfo{person}{Yuan Liu}, \bibinfo{person}{Christian Theobalt},
  \bibinfo{person}{Taku Komura}, {and} \bibinfo{person}{Wenping Wang}.}
  \bibinfo{year}{2021}\natexlab{}.
\newblock \showarticletitle{NeuS: Learning Neural Implicit Surfaces by Volume
  Rendering for Multi-view Reconstruction}. In
  \bibinfo{booktitle}{\emph{Advances in Neural Information Processing Systems
  34: Annual Conference on Neural Information Processing Systems 2021, NeurIPS
  2021, December 6-14, 2021, virtual}}. \bibinfo{pages}{27171--27183}.
\newblock
\urldef\tempurl%
\url{https://proceedings.neurips.cc/paper/2021/hash/e41e164f7485ec4a28741a2d0ea41c74-Abstract.html}
\showURL{%
\tempurl}


\bibitem[Wizadwongsa et~al\mbox{.}(2021)]%
        {nex}
\bibfield{author}{\bibinfo{person}{Suttisak Wizadwongsa},
  \bibinfo{person}{Pakkapon Phongthawee}, \bibinfo{person}{Jiraphon
  Yenphraphai}, {and} \bibinfo{person}{Supasorn Suwajanakorn}.}
  \bibinfo{year}{2021}\natexlab{}.
\newblock \showarticletitle{NeX: Real-Time View Synthesis With Neural Basis
  Expansion}. In \bibinfo{booktitle}{\emph{{IEEE} Conference on Computer Vision
  and Pattern Recognition, {CVPR} 2021, virtual, June 19-25, 2021}}.
  \bibinfo{publisher}{Computer Vision Foundation / {IEEE}},
  \bibinfo{pages}{8534--8543}.
\newblock
\urldef\tempurl%
\url{https://doi.org/10.1109/CVPR46437.2021.00843}
\showDOI{\tempurl}


\bibitem[Yang et~al\mbox{.}(2022)]%
        {neumesh}
\bibfield{author}{\bibinfo{person}{Bangbang Yang}, \bibinfo{person}{Chong Bao},
  \bibinfo{person}{Junyi Zeng}, \bibinfo{person}{Hujun Bao},
  \bibinfo{person}{Yinda Zhang}, \bibinfo{person}{Zhaopeng Cui}, {and}
  \bibinfo{person}{Guofeng Zhang}.} \bibinfo{year}{2022}\natexlab{}.
\newblock \showarticletitle{NeuMesh: Learning Disentangled Neural Mesh-based
  Implicit Field for Geometry and Texture Editing}.
\newblock \bibinfo{journal}{\emph{CoRR}}  \bibinfo{volume}{abs/2207.11911}
  (\bibinfo{year}{2022}).
\newblock


\bibitem[Yariv et~al\mbox{.}(2021)]%
        {volsdf}
\bibfield{author}{\bibinfo{person}{Lior Yariv}, \bibinfo{person}{Jiatao Gu},
  \bibinfo{person}{Yoni Kasten}, {and} \bibinfo{person}{Yaron Lipman}.}
  \bibinfo{year}{2021}\natexlab{}.
\newblock \showarticletitle{Volume Rendering of Neural Implicit Surfaces}. In
  \bibinfo{booktitle}{\emph{Advances in Neural Information Processing Systems
  34: Annual Conference on Neural Information Processing Systems 2021, NeurIPS
  2021, December 6-14, 2021, virtual}},
  \bibfield{editor}{\bibinfo{person}{Marc'Aurelio Ranzato},
  \bibinfo{person}{Alina Beygelzimer}, \bibinfo{person}{Yann~N. Dauphin},
  \bibinfo{person}{Percy Liang}, {and} \bibinfo{person}{Jennifer~Wortman
  Vaughan}} (Eds.). \bibinfo{pages}{4805--4815}.
\newblock
\urldef\tempurl%
\url{https://proceedings.neurips.cc/paper/2021/hash/25e2a30f44898b9f3e978b1786dcd85c-Abstract.html}
\showURL{%
\tempurl}


\bibitem[Yariv et~al\mbox{.}(2020)]%
        {idr}
\bibfield{author}{\bibinfo{person}{Lior Yariv}, \bibinfo{person}{Yoni Kasten},
  \bibinfo{person}{Dror Moran}, \bibinfo{person}{Meirav Galun},
  \bibinfo{person}{Matan Atzmon}, \bibinfo{person}{Ronen Basri}, {and}
  \bibinfo{person}{Yaron Lipman}.} \bibinfo{year}{2020}\natexlab{}.
\newblock \showarticletitle{Multiview Neural Surface Reconstruction by
  Disentangling Geometry and Appearance}. In \bibinfo{booktitle}{\emph{Advances
  in Neural Information Processing Systems 33: Annual Conference on Neural
  Information Processing Systems 2020, NeurIPS 2020, December 6-12, 2020,
  virtual}}, \bibfield{editor}{\bibinfo{person}{Hugo Larochelle},
  \bibinfo{person}{Marc'Aurelio Ranzato}, \bibinfo{person}{Raia Hadsell},
  \bibinfo{person}{Maria{-}Florina Balcan}, {and} \bibinfo{person}{Hsuan{-}Tien
  Lin}} (Eds.).
\newblock
\urldef\tempurl%
\url{https://proceedings.neurips.cc/paper/2020/hash/1a77befc3b608d6ed363567685f70e1e-Abstract.html}
\showURL{%
\tempurl}


\bibitem[Yu et~al\mbox{.}(2021)]%
        {plenoctrees}
\bibfield{author}{\bibinfo{person}{Alex Yu}, \bibinfo{person}{Ruilong Li},
  \bibinfo{person}{Matthew Tancik}, \bibinfo{person}{Hao Li},
  \bibinfo{person}{Ren Ng}, {and} \bibinfo{person}{Angjoo Kanazawa}.}
  \bibinfo{year}{2021}\natexlab{}.
\newblock \showarticletitle{PlenOctrees for Real-time Rendering of Neural
  Radiance Fields}. In \bibinfo{booktitle}{\emph{{ICCV}}}.
  \bibinfo{publisher}{{IEEE}}, \bibinfo{pages}{5732--5741}.
\newblock


\bibitem[Yuan et~al\mbox{.}(2022)]%
        {nerf-editing}
\bibfield{author}{\bibinfo{person}{Yu{-}Jie Yuan}, \bibinfo{person}{Yang{-}Tian
  Sun}, \bibinfo{person}{Yu{-}Kun Lai}, \bibinfo{person}{Yuewen Ma},
  \bibinfo{person}{Rongfei Jia}, {and} \bibinfo{person}{Lin Gao}.}
  \bibinfo{year}{2022}\natexlab{}.
\newblock \showarticletitle{NeRF-Editing: Geometry Editing of Neural Radiance
  Fields}.
\newblock \bibinfo{journal}{\emph{CoRR}}  \bibinfo{volume}{abs/2205.04978}
  (\bibinfo{year}{2022}).
\newblock


\bibitem[Zhang et~al\mbox{.}(2021)]%
        {physg}
\bibfield{author}{\bibinfo{person}{Kai Zhang}, \bibinfo{person}{Fujun Luan},
  \bibinfo{person}{Qianqian Wang}, \bibinfo{person}{Kavita Bala}, {and}
  \bibinfo{person}{Noah Snavely}.} \bibinfo{year}{2021}\natexlab{}.
\newblock \showarticletitle{PhySG: Inverse Rendering With Spherical Gaussians
  for Physics-Based Material Editing and Relighting}. In
  \bibinfo{booktitle}{\emph{{IEEE} Conference on Computer Vision and Pattern
  Recognition, {CVPR} 2021, virtual, June 19-25, 2021}}.
  \bibinfo{publisher}{Computer Vision Foundation / {IEEE}},
  \bibinfo{pages}{5453--5462}.
\newblock
\urldef\tempurl%
\url{https://doi.org/10.1109/CVPR46437.2021.00541}
\showDOI{\tempurl}


\bibitem[Zhang et~al\mbox{.}(2020)]%
        {nerf++}
\bibfield{author}{\bibinfo{person}{Kai Zhang}, \bibinfo{person}{Gernot
  Riegler}, \bibinfo{person}{Noah Snavely}, {and} \bibinfo{person}{Vladlen
  Koltun}.} \bibinfo{year}{2020}\natexlab{}.
\newblock \showarticletitle{NeRF++: Analyzing and Improving Neural Radiance
  Fields}.
\newblock \bibinfo{journal}{\emph{CoRR}}  \bibinfo{volume}{abs/2010.07492}
  (\bibinfo{year}{2020}).
\newblock
\showeprint[arXiv]{2010.07492}
\urldef\tempurl%
\url{https://arxiv.org/abs/2010.07492}
\showURL{%
\tempurl}


\bibitem[Zielonka et~al\mbox{.}(2022)]%
        {instant-volumetric-head-avatars}
\bibfield{author}{\bibinfo{person}{Wojciech Zielonka}, \bibinfo{person}{Timo
  Bolkart}, {and} \bibinfo{person}{Justus Thies}.}
  \bibinfo{year}{2022}\natexlab{}.
\newblock \showarticletitle{Instant Volumetric Head Avatars}.
\newblock \bibinfo{journal}{\emph{CoRR}}  \bibinfo{volume}{abs/2211.12499}
  (\bibinfo{year}{2022}).
\newblock
\urldef\tempurl%
\url{https://doi.org/10.48550/arXiv.2211.12499}
\showDOI{\tempurl}
\showeprint[arXiv]{2211.12499}


\end{thebibliography}





\end{document}